\documentclass[11pt]{article}

\usepackage[preprint]{acl}
\usepackage{times}
\usepackage{latexsym}
\usepackage{xurl}
\usepackage[T1]{fontenc}
\usepackage[utf8]{inputenc}
\usepackage{microtype}
\usepackage{inconsolata}
\usepackage{graphicx}
\usepackage{subcaption}
\usepackage{booktabs}
\usepackage{bm}
\usepackage{amsmath}
\usepackage{multirow}
\usepackage{xcolor}
\usepackage{colortbl}
\usepackage{amssymb} 
\usepackage{makecell}
\usepackage[dvipsnames]{xcolor}  
\usepackage{tocloft}
\usepackage{hyperref}
\usepackage{float}
\newcommand*\samethanks[1][\value{footnote}]{\footnotemark[#1]}
\hypersetup{
    colorlinks=true,
    filecolor=magenta,
    urlcolor=cyan,
}

\newcommand{\cf}{\emph{cf. }}

\newcommand{\ie}{\emph{i.e., }}

\definecolor{color+}{RGB}{0, 100, 0}
\definecolor{color-}{RGB}{200, 0, 0}
\definecolor{lightgray}{gray}{0.9}
\definecolor{lightred}{RGB}{255, 230, 230}
\definecolor{lightgreen}{rgb}{0.9, 1, 0.9}

\title{Understanding Multilingualism in Mixture-of-Experts LLMs: \\ Routing Mechanism, Expert Specialization, and Layerwise Steering}
\author{Yuxin Chen$^{1}$\thanks{\ \ \ Equal contribution}, Zhengzhou Cai$^{2}$\samethanks, Xiangtian Ji$^1$, Weixiang Zhao$^3$,\\ \textbf{An Zhang}$^4$\thanks{\ \ Corresponding author} \textbf{Xiang Wang}$^4$, \textbf{Tat-Seng Chua}$^1$ \\
        $^1$National University of Singapore, $^2$Beijing University of Posts and Telecommunication, \\
        $^3$Harbin Institute of Technology, $^4$University of Science and Technology of China\\
        \texttt{yuxin.chen@u.nus.edu}, }
\begin{document}
    \maketitle

\begin{abstract}

Mixture-of-Experts (MoE) architectures have shown strong multilingual capabilities, yet the internal mechanisms underlying performance gains and cross-language differences remain insufficiently understood.
In this work, we conduct a systematic analysis of MoE models, examining routing behavior and expert specialization across languages and network depth.
Our analysis reveals that multilingual processing in MoE models is highly structured: routing aligns with linguistic families, expert utilization follows a clear layerwise pattern, and high-resource languages rely on shared experts while low-resource languages depend more on language-exclusive experts despite weaker performance.
Layerwise interventions further show that early and late MoE layers support language-specific processing, whereas middle layers serve as language-agnostic capacity hubs.
Building on these insights, we propose a routing-guided steering method that adaptively guides routing behavior in middle layers toward shared experts associated with dominant languages at inference time, leading to consistent multilingual performance improvements, particularly for linguistically related language pairs. Our code is available at \url{https://github.com/conctsai/Multilingualism-in-Mixture-of-Experts-LLMs}.
\end{abstract}
\addtocontents{toc}{\protect\setcounter{tocdepth}{-1}}
\section{Introduction}
\label{sec:introduction}

The Mixture-of-Experts (MoE) architecture, comprising sparse expert networks and a routing mechanism, has emerged as a prominent design for scaling large language models (LLMs)~\citep{gpt-oss, Deepseek-R1}.
By conditionally activating a subset of parameters, MoE models enable substantial parameter scaling with manageable training and inference costs under comparable compute budgets~\citep{moe-2017}.
Consequently, recent MoE-based LLMs have demonstrated strong multilingual capabilities compared with their dense counterparts, benefiting from sparse activation of specialized experts through routing~\citep{deepseekmoe}.
Despite these gains, clear performance disparities persist across languages with different resource levels~\citep{Qwen3}, raising fundamental questions about how multilingual capacity is utilized within MoE models.
Understanding the internal mechanisms that give rise to such disparities is essential for both interpretability and targeted improvement.

From an interpretability perspective, however, the current understanding of multilingual capabilities in LLMs has been largely developed in the context of dense architectures~\cite{yiran_howdo}.
Prior work has extensively investigated how multilingual representations emerge in dense models, through analyses of internal activations~\citep{activation_1, activation_2}, shared neurons~\citep{AbstractThought}, or attention patterns across languages~\citep{attention_1, attention_2, attention_3}.
This line of work relies on the assumption that model parameters are shared across all inputs, which fundamentally differs from the sparse and specialized computation paradigm of MoE, where routing mechanisms and expert specialization enable different tokens and languages to activate distinct subsets of parameters~\citep{moe-1991, moe-survey}.
As a result, how MoE architecture contribute to multilingual behavior, and how such contributions vary across languages, remains insufficiently understood~\citep{deepseek-v3}.

In this work, we conduct a systematic analysis of multilingual MoE models, focusing on routing behavior and expert specialization across languages and network depth.
Our analysis reveals that multilingual behavior in MoE models is highly structured.
Across languages, routing behavior exhibits clear regularities: languages within the same linguistic family tend to share similar routing distributions, whereas linguistically distant languages are routed through more distinct subsets of experts (\cf Section \ref{subsec:routing_findings}).
Moreover, both routing similarity and expert utilization display a pronounced layerwise structure.
Middle layers are characterized by higher cross-language routing similarity and a predominant reliance on shared experts, while early and late layers exhibit stronger language-specific routing and a higher concentration of language-exclusive experts (\cf Section \ref{subsec:routing_findings} \& \ref{subsec:expert_findings}).
This structural pattern is further modulated by language resource levels.
Dominant languages serve as central hubs for cross-lingual capacity sharing, high-resource languages rely heavily on shared experts, whereas low-resource languages depend more on language-exclusive experts yet remain weak (\cf Section \ref{subsec:expert_findings}).
Together, these observations point to a functional stratification of multilingual processing in MoE models, in which different MoE layers play distinct roles in language understanding, cross-lingual transfer, and language-specific generation.

To causally validate this hypothesis, we conduct a layerwise intervention study by selectively masking language-exclusive experts during inference (\cf Section \ref{subsec:intervention_method}).
The results reveal distinct functional roles across depth: masking early-layer exclusive experts severely degrades language understanding, masking late-layer exclusive experts disrupts language consistency during generation, while interventions in middle layers have minimal impact on performance.
These findings indicate that middle layers function as language-agnostic capacity hubs, supporting cross-lingual knowledge sharing, whereas early and late layers are responsible for language-specific processing (\cf Section \ref{subsec:intervention_results}).

Building on this structured understanding, we further propose a routing-guided steering method to improve multilingual performance at inference time.
By adaptively guiding routing behavior in middle layers toward shared experts associated with dominant languages—while leaving early and late layers unchanged—we enable more effective cross-lingual capacity transfer without disrupting language-specific understanding or generation (\cf Section \ref{subsec:steering_method}).
Experimental results demonstrate consistent performance improvements across both high-resource and low-resource languages, with gains strongly correlated with linguistic proximity to the steering source language (\cf Section \ref{subsec:steering_results}).

\section{Preliminary} 
\label{sec:preliminary}

In this section, we formally summarize the token-level routing mechanism employed in MoE LLMs. 
Unlike dense models that apply the same parameters to all inputs, MoE architectures introduce sparsity by conditionally activating only a subset of parameters.

Consider an MoE model with $L$ layers. 
The $l$-th MoE layer consists of two core components: a set of $E$ expert networks denoted as $\{f_{l,1}, \dots, f_{l,E}\}$, where each expert is typically a feed-forward network (FFN), and a gating network $G_l$ that acts as a router to distribute input tokens to a selected subset of experts.

For a token $x$ in the input sequence, let $\mathbf{h}_l(x) \in \mathbb{R}^d$ denote its hidden representation at layer $l$. 
The router computes routing logits by projecting $\mathbf{h}_l(x)$ through a learnable weight matrix $\mathbf{W}_l \in \mathbb{R}^{d \times E}$:
\begin{equation}
\mathbf{g}_l(x) = \mathbf{W}_l \mathbf{h}_l(x) \in \mathbb{R}^{E}.
\end{equation}

To enforce sparsity, a Top-$K$ routing strategy selects the indices of $K$ experts with the highest logits, denoted as $\mathcal{S}_l(x)$.
The expert probabilities $\boldsymbol{\alpha}_l(x)$ are then computed by applying a softmax over these selected logits:
\begin{equation}
\boldsymbol{\alpha}_l(x) = \mathrm{softmax}\left ( \mathrm{Top}\text{-}K\!\left(\mathbf{g}_l(x)\right)  \right ).
\end{equation}

The output of the $l$-th MoE layer is computed as the linearly weighted sum of the activations from the selected experts:
\begin{equation}
\mathbf{h}_{l+1}(x) = \sum_{i \in \mathcal{S}_l(x)} \alpha_{l,i}(x) \cdot f_{l,i}(\mathbf{h}_l(x)).
\end{equation}

\section{Analysis Setup}
\label{sec:analysis_setup}
\paragraph{Model.}
We conduct our analysis on Qwen3-30B-A3B~\citep{Qwen3}, an advanced and representative MoE LLM.

\paragraph{Benchmark.}
To conduct analysis on multilingual processing, we adopt the \textsc{Belebele}~\citep{Belebele} dataset, a multilingual natural language understanding benchmark, covering a wide range of languages with minimal domain variation.

\paragraph{Languages.}
We mainly analyze 10 languages spanning diverse linguistic families and resource levels:
Arabic (Ar), Bengali (Bn), German (De), English (En), Spanish (Es), French (Fr), Japanese (Ja), Korean (Ko), Swahili (Sw), and Chinese (Zh).
English and Chinese are treated as dominant languages due to their prevalence in the training data~\cite{Qwen3}.
German, Spanish, French, Japanese, Korean, and Arabic are regarded as high-resource languages, while Swahili and Bengali represent low-resource languages~\cite{PolyMath, MGSM}.
In practice, the exact set of evaluated languages may vary across benchmarks, depending on their language coverage.

\section{Analysis of Routing Behavior}

As the defining component that differentiates MoE models from dense architectures, the router determines how input tokens are dynamically allocated to experts.
In this section, we analyze routing behavior, focusing on how expert selection patterns vary across linguistic inputs and network depth.

\subsection{Metrics}
\label{subsec:routing_metrics}

\paragraph{Routing distribution.}
We characterize routing behavior via expert selection frequencies, \ie  measuring how frequently each expert is selected for tokens from a given language at each layer.

Let $\mathcal{D}_{\ell}$ denote the set of tokens associated with language $\ell$.
For each layer $l$ and expert index $i \in \{1,\dots,E\}$,
we define the routing frequency of expert $i$ for language $\ell$ as:
\begin{equation}
p_{l,i}^{(\ell)}
=
\frac{1}{|\mathcal{D}_{\ell}|}
\sum_{x \in \mathcal{D}_{\ell}}
\mathbb{I}\!\left(i \in \mathcal{S}_l(x)\right),
\end{equation}
where $\mathcal{S}_l(x)$ is the set of experts selected by the Top-$K$ router for token $x$ at layer $l$,
as defined in Section~2.
We aggregate the routing frequencies over all experts into a routing distribution:
\begin{equation}
\label{eq:routing_distribution}
\mathbf{P}_l^{(\ell)}
=
\big[
p_{l,1}^{(\ell)}, \dots, p_{l,E}^{(\ell)}
\big]^{\top}.
\end{equation}
which summarizes how expert selections are allocated across experts for language $\ell$ at layer $l$.

\paragraph{Cross-language routing similarity.}
To compare routing behavior across languages,
we measure the similarity between their routing distributions.

Given two languages $\ell_1$ and $\ell_2$,
we compute the Jensen--Shannon divergence between their routing distributions at layer $l$:
\begin{equation}
\begin{split}
    \mathrm{JSD}_l(\ell_1, \ell_2)
    =\;&
    \frac{1}{2}
    \mathrm{KL}\!\left(
    \mathbf{P}_l^{(\ell_1)} \parallel \mathbf{M}
    \right) \\
    +\;&
    \frac{1}{2}
    \mathrm{KL}\!\left(
    \mathbf{P}_l^{(\ell_2)} \parallel \mathbf{M}
    \right),
\end{split}
\end{equation}
where
\begin{equation}
\mathbf{M}
=
\frac{1}{2}
\left(
\mathbf{P}_l^{(\ell_1)} + \mathbf{P}_l^{(\ell_2)}
\right).
\end{equation}
We define routing similarity as
\begin{equation}
\label{eq:routing_similarity}
\mathrm{Sim}_l(\ell_1, \ell_2)
=
1 - \mathrm{JSD}_l(\ell_1, \ell_2),
\end{equation}
where larger values indicate more similar routing behavior between the two languages at layer $l$.

\subsection{Findings}
\label{subsec:routing_findings}

\begin{figure}[t]
  \centering
  \vspace{-5pt}
  \includegraphics[width=0.95\columnwidth]{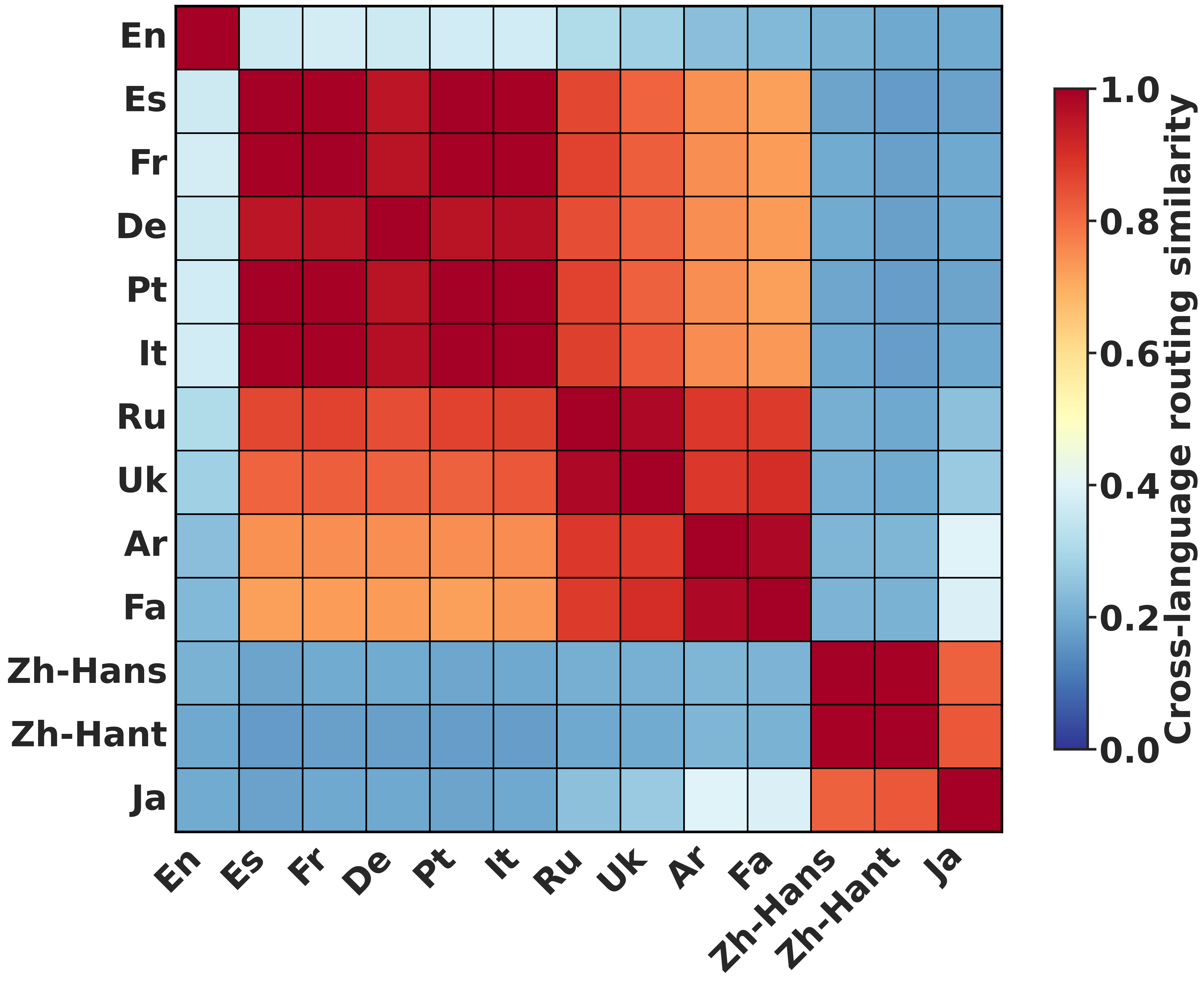}
  \vspace{-10pt}
  \caption{Pairwise cross-language routing similarity across different languages. Red indicates high similarity, while blue indicates high dissimilarity.
  \label{fig:js_divergence_heatmap}}
\end{figure}

\paragraph{Routing similarity reflects linguistic relatedness.}
Using the proposed cross-language routing similarity metric, we compare routing behaviors across languages.
Detailed experimental settings are provided in
Appendix~\ref{app:language family settings}.
Figure~\ref{fig:js_divergence_heatmap} reveals a clear block-wise structure in pairwise routing similarity.
Languages from the same linguistic family or share similar writing systems tend to exhibit higher routing similarity than linguistically distant languages.
This indicates that such languages tend to be routed through overlapping subsets of experts with comparable selection frequencies.
Such patterns persist across multiple language families, exemplified by Cyrillic-script languages such as Russian and Ukrainian, as well as CJK-related languages including Simplified Chinese, Traditional Chinese, and Japanese.
These observations suggest that the router organizes multilingual inputs in a family-structured manner, implicitly aligning routing behaviors among related languages.

\begin{figure*}[t]
  \centering
  \includegraphics[width=0.9\linewidth]{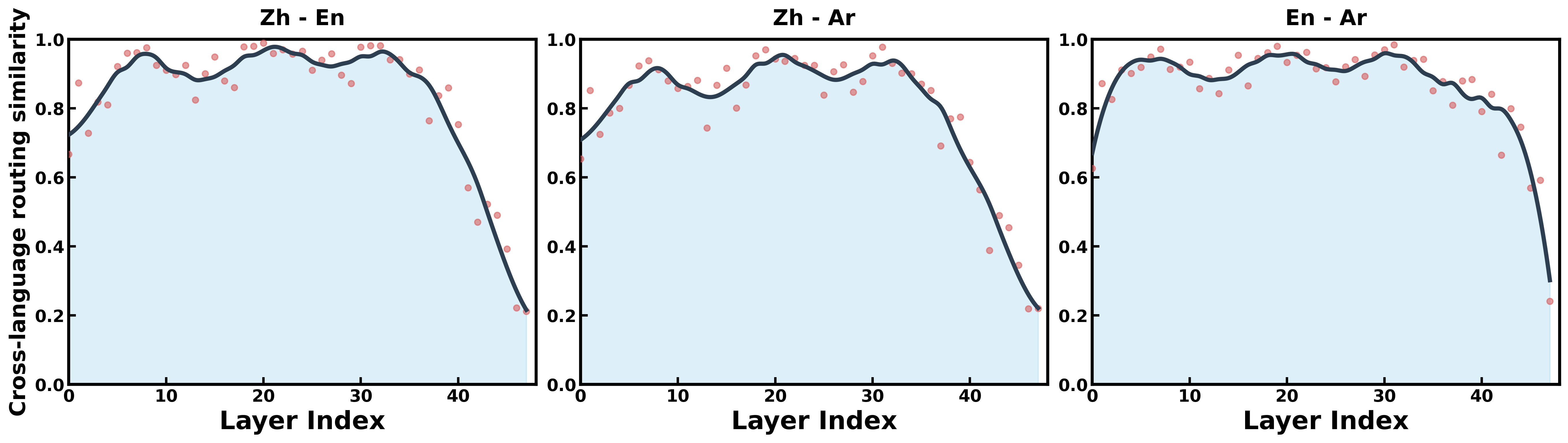}
  \vspace{-5pt}
  \caption{Layer-wise cross-language routing similarity.
The figure comprises three subplots representing pairwise combinations of three representative languages: English (En), Chinese (Zh) and Arabic (Ar).}
\vspace{-5pt}
  \label{fig:pairwise_layer_divergence}
\end{figure*}

\paragraph{Routing similarity exhibits layerwise structure.} 
Beyond differences across language families, routing similarity also exhibits a systematic depth-wise pattern.
Figure \ref{fig:pairwise_layer_divergence} shows that, across different language pairs, cross-language routing similarity is highest in middle layers and decreases toward both early and late layers.
Higher similarity corresponds to greater overlap in expert utilization across languages, whereas lower similarity reflects increasingly language-specific routing behavior.
This layerwise trend reveals a functional stratification of routing dynamics: routing decisions in middle layers are more language-agnostic, while early and late layers exhibit stronger language differetiation, aligning to recent study ~\cite{multilingual-moe}.
Together, these findings demonstrate that multilingual routing behavior in MoE models is structured not only across languages but also across network depth.
We provide additional analyses of routing entropy patterns and extended cross-lingual similarity results across a broader set of languages in Appendix~\ref{app:findings}.
\section{Analysis of Expert Specialization}
\label{sec:expert_analysis}
While routing analysis characterizes how the router distributes tokens across experts, expert analysis focuses on how individual experts are associated with different languages.
In this section, we analyze expert-level statistics, providing a complementary perspective to the routing-centric analysis.

\subsection{Metrics}
\label{subsec:expert_metrics}

\paragraph{Language-related experts.}
To identify experts that are most strongly associated with each language, we measure language-expert association using routing frequencies defined in Section~\ref{subsec:routing_metrics}.
For each layer $l$, language $\ell$, and expert index $i \in \{1,\dots,E\}$,
the routing frequency $p_{l,i}^{(\ell)}$ measures how often expert $i$
is selected for tokens from language $\ell$ at layer $l$.

Based on routing frequencies, we identify language-related experts by selecting the Top-$K$ experts with the highest values of $p_{l,i}^{(\ell)}$ at each layer for each language:
\begin{equation}
\label{eq:language-related-experts}
\mathcal{T}_l^{(\ell)}
=
\operatorname{Top\text{-}}K
\left(
\left\{ p_{l,i}^{(\ell)} \right\}_{i=1}^{E}
\right),
\end{equation}
where $K$ is a hyperparameter controlling the number of selected experts.
The language-related experts set $\mathcal{T}_l^{(\ell)}$ represents experts
that are most frequently selected for language $\ell$ at layer $l$.

\paragraph{Language-exclusive and shared experts.}
To distinguish language-specific specialization from cross-lingual sharing,
we examine how the routing frequency of each language-related expert
is distributed across languages.

Let $\mathcal{L}$ denote the set of languages under analysis.
For an expert $i$ that appears in $\mathcal{T}_l^{(\ell)}$ for at least one language,
we define its normalized routing frequency with respect to language $\ell$ as
$W_{l,i}^{(\ell)}
=
\frac{p_{l,i}^{(\ell)}}
{\sum_{\ell' \in \mathcal{L}} p_{l,i}^{(\ell')}}$.
An expert is regarded as language-exclusive for language $\ell$
if its routing frequency is dominated by that language:
\begin{equation}
\label{eq:language-exclusive-experts}
\mathcal{E}_{l,\mathrm{excl}}^{(\ell)}
=
\left\{
i \;\middle|\;
W_{l,i}^{(\ell)} > \theta
\right\},
\end{equation}
where threshold $\theta \in (0,1)$ controls how dominant a single language must be
for an expert to be considered language-exclusive.

The remaining language-related experts, whose routing frequency is not dominated by any single language, are treated as language-shared:
\begin{equation}
\label{eq:shared_experts}
\mathcal{E}_{l,\mathrm{shared}}
=
\left\{
i \;\middle|\;
\max_{\ell \in \mathcal{L}} W_{l,i}^{(\ell)} \le \theta
\right\}.
\end{equation}

\subsection{Findings}
\label{subsec:expert_findings}

\begin{figure}[t]
  \centering
  \includegraphics[width=0.95\columnwidth]{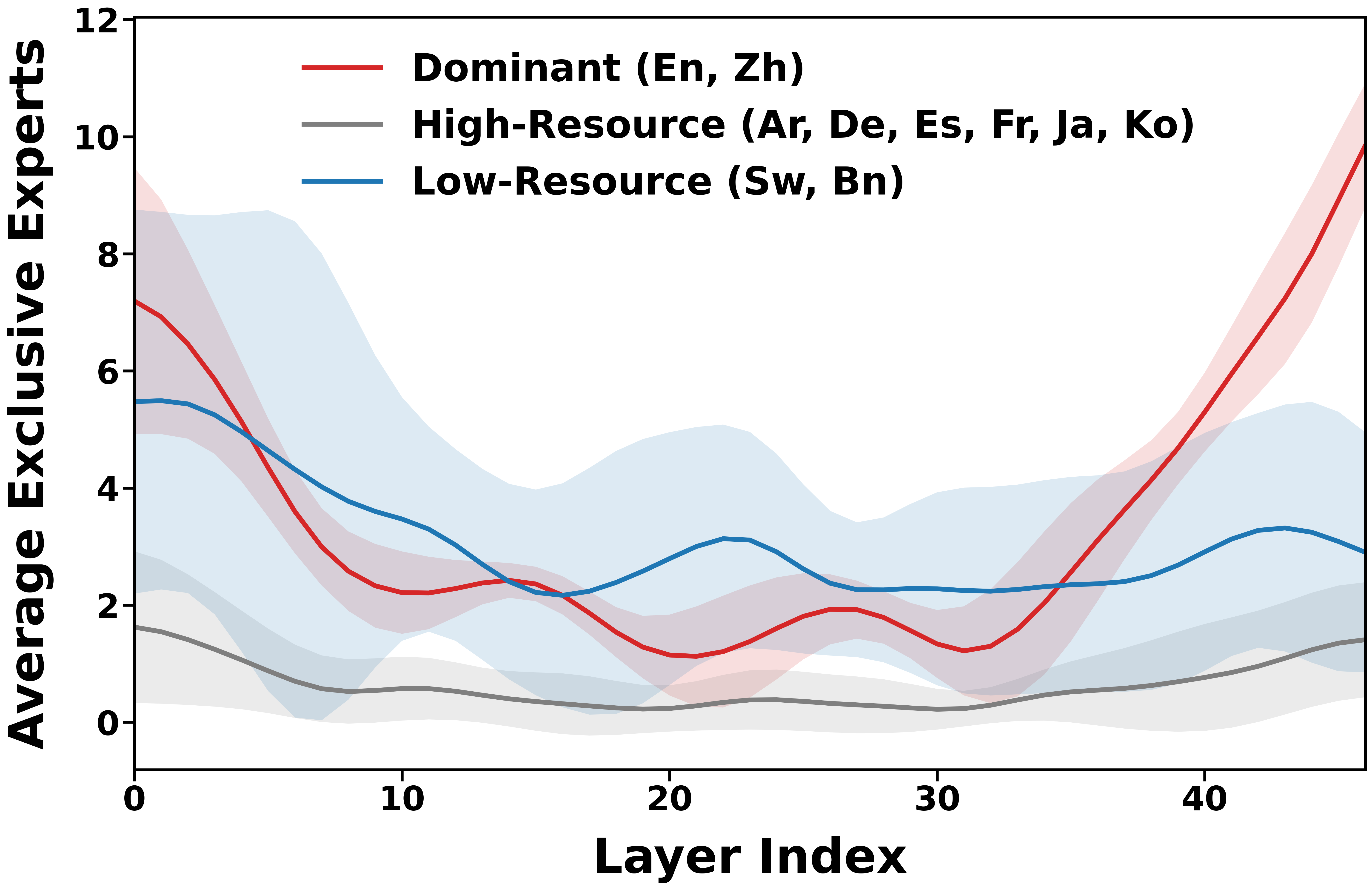}
  \vspace{-5pt}
  \caption{Average number of exclusive experts per layer for different language groups.
  \label{fig:layer_exclusive_experts}}
\end{figure}

\paragraph{Language-exclusive expert usage varies across language resource levels.}
Using the expert-level metrics defined in Section~\ref{subsec:expert_metrics}, we analyze how language-exclusive experts are distributed across languages with different resource levels.
The results are shown in Figure~\ref{fig:layer_exclusive_experts}, with detailed experimental settings provided in Appendix~\ref{app:expert specialization settings}.
Dominant languages, which constitute the primary source of pre-training data, are associated with the largest number of language-exclusive experts.
High-resource languages, in contrast, rely on remarkably few language-exclusive experts across the network, despite often achieving competitive performance.
Low-resource languages rely more heavily on language-exclusive experts, yet still lag behind in performance.

These contrasting patterns in expert utilization coincide with substantial differences in multilingual capability across languages with different resource levels \cite{Qwen3}.
Specifically, the strong performance of high-resource languages appears to be supported not by language-specific expert specialization, but by effective utilization of shared experts aligned with dominant languages, enabling cross-lingual capacity transfer.
In contrast, the heavier reliance on language-exclusive experts in low-resource languages suggests weaker integration into the shared expert space, which may limit access to transferable capacity and contribute to their persistent performance gap.

\paragraph{Layerwise distribution of language-exclusive experts.}
Beyond cross-language differences, we examine how language-exclusive experts
are distributed across network depth.
As shown in Figure~\ref{fig:layer_exclusive_experts}, the number of language-exclusive experts exhibits a consistent layerwise pattern across languages: it is relatively high in early layers, decreases substantially in middle layers, and increases again toward late layers, forming a clear U-shaped distribution.
This depth-wise pattern aligns with the routing analysis presented in the previous section.
Middle layers, which exhibit higher cross-language routing similarity,
are characterized by fewer language-exclusive experts, indicating more language-agnostic expert utilization.
In contrast, early and late layers contain a larger number of language-exclusive experts, reflecting stronger language-specific specialization.
Together, these observations suggest a functional stratification of expert utilization in MoE models, where language-agnostic processing predominantly occurs in middle layers, while language-specific specialization is concentrated in early and late layers.
\section{Layerwise Intervention Study}
\label{sec:intervention}

\begin{figure}[t]
  \centering
  \includegraphics[width=\linewidth]{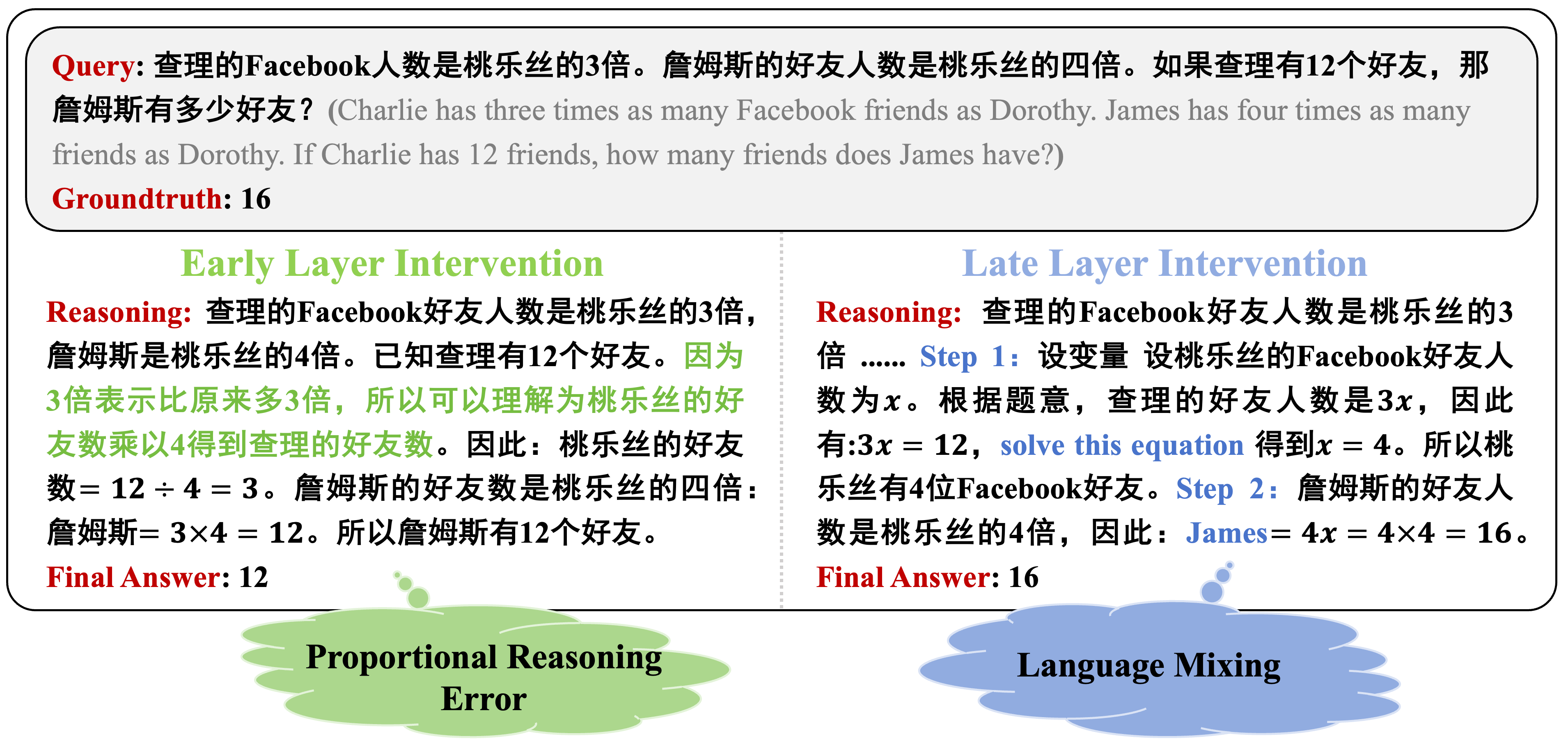}
  \vspace{-15pt}
  \caption{Case study of layerwise intervention on Chinese-exclusive experts. Early-layer intervention leads to understanding failure, while late-layer intervention results in language mixing.
  \label{fig:case_study}}
\end{figure}

\begin{table*}[t]
\centering
\small
\setlength{\tabcolsep}{4.5pt}
\vspace{-10pt}
\caption{
Layer-wise intervention effects on \textsc{MGSM} \textbf{accuracy}.
\textit{Target} reports the accuracy of a language when its own language-exclusive experts are masked, while \textit{Other (Avg.)} denotes the average accuracy of a language when the experts of other language are masked.
}
\renewcommand{\arraystretch}{1.3}

\begin{tabular}{c|cc|cccc|cc|c}
\toprule
\multirow{2}{*}{\makecell{\textbf{Intervened} \\ \textbf{Language}}} & 
\multicolumn{2}{c|}{\textbf{Low-resource}} & 
\multicolumn{4}{c|}{\textbf{High-resource}} & 
\multicolumn{2}{c|}{\textbf{Dominant}} & 
\textbf{Overall} \\
& Bn & Sw & De & Ja & Es & Fr & Zh & En & \textbf{Avg.} \\
\midrule

\rowcolor{lightgray}
- & 84.8 & 54.8 & 88.8 & 85.6 & 92.8 & 86.4 & 90.0 & 92.8 & 84.5 \\
\midrule

\multicolumn{10}{c}{\textbf{\textit{Intervention Layer: Early}}} \\
\midrule
\textit{Target} & 
53.6$^{\color{color-}-31.2}$ & 
6.8$^{\color{color-}-48.0}$ & 
85.6$^{\color{color-}-3.2}$ & 
82.8$^{\color{color-}-2.8}$ & 
92.4$^{\color{color-}-0.4}$ & 
87.2$^{\color{color+}+0.8}$ & 
84.4$^{\color{color-}-5.6}$ & 
92.0$^{\color{color-}-0.8}$ & 
\cellcolor{lightred}73.1$^{\color{color-}-11.4}$ \\

\textit{Other (Avg.)} & 
81.8$^{\color{color-}-2.7}$ & 
87.1$^{\color{color-}-1.6}$ & 
82.6$^{\color{color-}-1.3}$ & 
83.9$^{\color{color-}-0.4}$ & 
82.5$^{\color{color-}-0.8}$ & 
83.7$^{\color{color-}-0.5}$ & 
82.4$^{\color{color-}-1.3}$ & 
81.8$^{\color{color-}-1.5}$ & 
83.3$^{\color{color-}-1.2}$ \\
\midrule

\multicolumn{10}{c}{\textbf{\textit{Intervention Layer: Middle}}} \\
\midrule
\textit{Target} & 
84.0$^{\color{color-}-0.8}$ & 
48.0$^{\color{color-}-6.8}$ & 
90.0$^{\color{color+}+1.2}$ & 
86.0$^{\color{color+}+0.4}$ & 
92.8$^{\color{black}+0.0}$ & 
86.4$^{\color{black}+0.0}$ & 
90.0$^{\color{black}+0.0}$ & 
92.0$^{\color{color-}-0.8}$ & 
83.6$^{\color{color-}-0.9}$ \\

\textit{Other (Avg.)} & 
83.4$^{\color{color-}-1.1}$ & 
88.9$^{\color{color+}+0.2}$ & 
84.0$^{\color{color+}+0.1}$ & 
84.0$^{\color{color-}-0.3}$ & 
83.0$^{\color{color-}-0.3}$ & 
83.5$^{\color{color-}-0.7}$ & 
83.9$^{\color{color+}+0.2}$ & 
83.1$^{\color{color-}-0.2}$ & 
84.2$^{\color{color-}-0.3}$ \\
\midrule

\multicolumn{10}{c}{\textbf{\textit{Intervention Layer: Late}}} \\
\midrule
\textit{Target} & 
72.4$^{\color{color-}-12.4}$ & 
20.4$^{\color{color-}-34.4}$ & 
90.4$^{\color{color+}+1.6}$ & 
82.8$^{\color{color-}-2.8}$ & 
91.2$^{\color{color-}-1.6}$ & 
88.0$^{\color{color+}+1.6}$ & 
87.6$^{\color{color-}-2.4}$ & 
94.4$^{\color{color+}+1.6}$ & 
\cellcolor{lightred}78.4$^{\color{color-}-6.1}$ \\

\textit{Other (Avg.)} & 
83.1$^{\color{color-}-1.4}$ & 
90.1$^{\color{color+}+1.4}$ & 
84.1$^{\color{color+}+0.2}$ & 
83.6$^{\color{color-}-0.7}$ & 
83.4$^{\color{color+}+0.1}$ & 
83.6$^{\color{color-}-0.6}$ & 
83.6$^{\color{color-}-0.1}$ & 
82.0$^{\color{color-}-1.3}$ & 
84.2$^{\color{color-}-0.3}$ \\
\bottomrule
\end{tabular}

\label{tab:intervention_mgsm}
\end{table*}

\begin{table*}[t]
\centering
\small
\setlength{\tabcolsep}{4.3pt}
\caption{
Layer-wise intervention effects on \textbf{language consistency} in \textsc{MGSM}.
\textit{Target} reports the consistency score of a language when its own language-exclusive experts are masked, while \textit{Other (Avg.)} denotes the average consistency of a language when the experts of other languages are masked.
}
\renewcommand{\arraystretch}{1.3}

\begin{tabular}{c|cc|cccc|cc|c}
\toprule
\multirow{2}{*}{\makecell{\textbf{Intervened} \\ \textbf{Language}}} & 
\multicolumn{2}{c|}{\textbf{Low-resource}} & 
\multicolumn{4}{c|}{\textbf{High-resource}} & 
\multicolumn{2}{c|}{\textbf{Dominant}} & 
\textbf{Overall} \\
& Bn & Sw & De & Ja & Es & Fr & Zh & En & \textbf{Avg.} \\
\midrule

\rowcolor{lightgray}
- & 97.6 & 90.8 & 86.0 & 74.8 & 95.6 & 88.0 & 95.2 & 98.8 & 90.9 \\
\midrule

\multicolumn{10}{c}{\textbf{\textit{Intervention Layer: Early}}} \\
\midrule
\textit{Target} & 
99.2$^{\color{color+}+1.6}$ & 
95.6$^{\color{color+}+4.8}$ & 
86.8$^{\color{color+}+0.8}$ & 
75.2$^{\color{color+}+0.4}$ & 
96.0$^{\color{color+}+0.4}$ & 
89.6$^{\color{color+}+1.6}$ & 
94.4$^{\color{color-}-0.8}$ & 
99.2$^{\color{color+}+0.4}$ & 
92.0$^{\color{color+}+1.1}$ \\

\textit{Other (Avg.)} & 
97.8$^{\color{color+}+0.2}$ & 
90.9$^{\color{color+}+0.1}$ & 
86.1$^{\color{color+}+0.1}$ & 
75.0$^{\color{color+}+0.2}$ & 
95.5$^{\color{color-}-0.1}$ & 
87.9$^{\color{color-}-0.1}$ & 
95.3$^{\color{color+}+0.1}$ & 
98.7$^{\color{color-}-0.1}$ & 
90.9$^{\color{black}+0.0}$ \\
\midrule

\multicolumn{10}{c}{\textbf{\textit{Intervention Layer: Middle}}} \\
\midrule
\textit{Target} & 
98.0$^{\color{color+}+0.4}$ & 
88.8$^{\color{color-}-2.0}$ & 
86.0$^{\color{black}+0.0}$ & 
74.4$^{\color{color-}-0.4}$ & 
94.4$^{\color{color-}-1.2}$ & 
88.4$^{\color{color+}+0.4}$ & 
95.2$^{\color{black}+0.0}$ & 
99.6$^{\color{color+}+0.8}$ & 
90.6$^{\color{color-}-0.3}$ \\

\textit{Other (Avg.)} & 
97.5$^{\color{color-}-0.1}$ & 
91.0$^{\color{color+}+0.2}$ & 
86.2$^{\color{color+}+0.2}$ & 
74.7$^{\color{color-}-0.1}$ & 
95.5$^{\color{color-}-0.1}$ & 
88.1$^{\color{color+}+0.1}$ & 
95.3$^{\color{color+}+0.1}$ & 
98.7$^{\color{color-}-0.1}$ & 
90.9$^{\color{black}+0.0}$ \\
\midrule

\multicolumn{10}{c}{\textbf{\textit{Intervention Layer: Late}}} \\
\midrule
\textit{Target} & 
24.8$^{\color{color-}-72.8}$ & 
85.6$^{\color{color-}-5.2}$ & 
12.4$^{\color{color-}-73.6}$ & 
50.8$^{\color{color-}-24.0}$ & 
90.4$^{\color{color-}-5.2}$ & 
3.6$^{\color{color-}-84.4}$ & 
89.2$^{\color{color-}-6.0}$ & 
98.8$^{\color{black}+0.0}$ & 
\cellcolor{lightred}56.9$^{\color{color-}-33.9}$ \\

\textit{Other (Avg.)} & 
97.5$^{\color{color-}-0.1}$ & 
90.7$^{\color{color-}-0.1}$ & 
86.2$^{\color{color+}+0.2}$ & 
74.9$^{\color{color+}+0.1}$ & 
95.6$^{\color{black}+0.0}$ & 
87.9$^{\color{color-}-0.1}$ & 
95.3$^{\color{color+}+0.1}$ & 
98.7$^{\color{color-}-0.1}$ & 
90.8$^{\color{color-}-0.1}$ \\
\bottomrule
\end{tabular}

\label{tab:intervention_mgsm_consistency}
\end{table*}

\begin{table}[t]
\centering
\small
\setlength{\tabcolsep}{4.5pt}
\renewcommand{\arraystretch}{1.3}
\caption{
Layer-wise intervention effects on \textbf{language consistency} in \textsc{FLORES-200}.
}
\resizebox{\columnwidth}{!}{
\begin{tabular}{c|cc|cccc|cc|c}
\toprule
\multirow{2}{*}{\textbf{Layers}} &
\multicolumn{2}{c|}{\textbf{Low}} &
\multicolumn{4}{c|}{\textbf{High}} &
\multicolumn{2}{c|}{\textbf{Dominant}} &
\textbf{Overall} \\
& Bn & Sw & De & Ja & Es & Fr & Zh & En & \textbf{Avg.} \\
\midrule

\rowcolor{lightgray}
- & 99.6 & 99.1 & 95.7 & 99.6 & 93.6 & 90.8 & 99.4 & 99.7 & 97.2 \\

\textit{Early} & 99.9 & 97.5 & 95.6 & 99.4 & 93.4 & 90.4 & 99.3 & 99.7 & 96.9 \\

\textit{Middle} & 99.6 & 98.9 & 95.9 & 99.7 & 93.5 & 90.2 & 99.5 & 99.7 & 97.1 \\

\textit{Late} & 46.4 & 82.1 & 21.3 & 73.7 & 85.6 & 13.1 & 65.7 & 99.7 & \cellcolor{lightred}60.9 \\

\bottomrule
\end{tabular}
}

\label{tab:intervention_flores_compact}
\end{table}

Motivated by the functional stratification hypothesis in Section~\ref{subsec:expert_findings}, we conduct a layerwise intervention study to causally validate and further elucidate this hypothesis.
Specifically, by masking language-exclusive experts at different depths during inference, we aim to verify their distinct functional roles and characterize how early, middle, and late MoE layers contribute to multilingual processing.

\subsection{Intervention Method}
\label{subsec:intervention_method}
For a target language $\ell$, we selectively mask experts identified as language-exclusive for $\ell$
while leaving all other experts unchanged.
Given the router logits $\mathbf{g}_l(x) \in \mathbb{R}^E$ for token $x$ at layer $l$,
we define a masking set $\mathcal{M}_l^{(\ell)}$ that contains experts
identified as language-exclusive for language $\ell$ at layer $l$,
i.e., $\mathcal{M}_l^{(\ell)} = \mathcal{E}_{l,\mathrm{excl}}^{(\ell)}$.
For experts in $\mathcal{M}_l^{(\ell)}$, their logits are replaced with a large negative constant:
\begin{equation}
\tilde{g}_{l,i}(x) =
\begin{cases}
\nu, & \text{if } i \in \mathcal{M}_l^{(\ell)}, \\
g_{l,i}(x), & \text{otherwise},
\end{cases}
\end{equation}
where $\nu = -10^9$, effectively preventing these experts from being selected by the router.
To study depth-wise effects, we apply this intervention separately to early, middle, and late layers.
Detailed experimental settings for the expert intervention study are provided in
Appendix~\ref{app:expert intervention settings}.

\subsection{Intervention Results}
\label{subsec:intervention_results}

To assess the functional impact of masking language-exclusive experts, we conduct a comprehensive evaluation covering both reasoning correctness and language consistency.
Specifically, we evaluate reasoning performance on MGSM~\citep{MGSM}, a multilingual math reasoning benchmark, and measure multilingual understanding ability on XQuAD~\citep{xquad}, a multilingual benchmark focused on language understanding.
We evaluate language consistency on MGSM, and adopt FLORES-200~\citep{flore200} as a strictly language-conditioned benchmark, where the model is explicitly instructed to generate text in a specified target language.

\paragraph{Reasoning Performance.}
Table~\ref{tab:intervention_mgsm} summarizes the impact of masking language-exclusive experts at different depths on MGSM.
Across all settings, we observe a clear depth-dependent effect that closely aligns with the routing patterns identified in Section~\ref{subsec:expert_findings}.
Masking language-exclusive experts in early or late layers causes a clear accuracy drop for the corresponding target language.
Notably, intervening on experts that are exclusive to other languages has little effect on the target language’s performance, supporting the language-specific association of these exclusive experts.
We also notice that early-layer masking leads to substantially larger degradations than late-layer masking.
In contrast, masking language-exclusive experts in the middle layers yields negligible changes in accuracy across languages, providing causal evidence that the middle layers operate in a language-agnostic manner, with shared experts playing a central role.
To better understand the sources of performance degradation, we further analyze the model’s generated responses under early- and late-layer interventions.
Additional analyses on the relationship between language and reasoning performance are provided in Appendix~\ref{app:robustness}.

\paragraph{Effects of Early-Layer Interventions.}
When language-exclusive experts in early layers are masked, we observe pronounced failures in query understanding.
The model frequently confuses during problem analysis, repeatedly misinterpreting or distorting problem conditions, which in turn leads to erroneous reasoning trajectories and difficulty progressing toward a valid solution.
Representative examples are shown in Figure~\ref{fig:case_study}.
Consistent with our qualitative observations, masking early-layer language-exclusive experts leads to a substantial performance degradation on XQuAD (\cf Table \ref{tab:xquad_intervention}), confirming that early-layer exclusive experts are essential for multilingual understanding.

\paragraph{Effects of Late-Layer Interventions.}
Masking language-exclusive experts in late layers leads to a different pattern.
In this setting, overall task accuracy is often mildly affected, and for some high-resource and dominant languages, performance can even improve slightly.
However, despite relatively preserved accuracy, the final outputs exhibit severe language mixing (see Figure~\ref{fig:case_study}).

This phenomenon is quantitatively confirmed by the language consistency results.
As shown in Table~\ref{tab:intervention_mgsm_consistency}, masking late-layer language-exclusive experts causes a dramatic drop in language consistency for the corresponding target language, while the consistency remains largely unaffected when masking experts of other languages.
Notably, this degradation occurs even when the task itself remains solvable, revealing that language identity control and reasoning competence are not intrinsically coupled.
We further validate this effect under a strictly language-conditioned generation setting using FLORES-200.
As shown in Table~\ref{tab:intervention_flores_compact}, late-layer interventions lead to a severe collapse in translation language consistency across most non-dominant languages.
Remarkably, this degradation persists even when the source and target languages are identical, i.e., when the model is simply asked to reproduce text in the same language.
Taken together, these results provide causal evidence that late-layer language-exclusive experts are essential to reliably produce outputs in the intended target language.
Per-language results are reported in Appendix~\ref{app:addition-expert-intervention}, and additional analyses are provided in Appendix~\ref{app:Analysis}.

\section{Routing Steering for Multilingual Enhancement}
\label{sec:enhancement}

\begin{table*}[t]
\centering
\renewcommand{\arraystretch}{1.3}
\setlength{\tabcolsep}{4pt}
\caption{
Performance comparison on the PolyMath.
Steering vectors are constructed from two dominant source languages (En and Zh) and applied to different target languages.
}
\small
\begin{tabular}{c|cccccc|c}
\toprule
\multirow{2}{*}{\textbf{Steering Source}} & \multicolumn{6}{c|}{\textbf{Target Language}} & \multirow{2}{*}{\textbf{Avg}} \\
 & {Bn} & {Sw} & {Fr} & {Es} & {Ja} & {De} & \\
\midrule

\rowcolor{lightgray}
- 
& 85.6 & 52.8 & 86.4 & 92.8 & 82.4 & 84.8 & 80.8 \\

\midrule
\textit{En}
& 86.4$^{\color{color+}+0.8}$ 
& 60.0$^{\color{color+}+7.2}$ 
& 88.0$^{\color{color+}+1.6}$ 
& 94.4$^{\color{color+}+1.6}$ 
& 82.4$^{\color{black}+0.0}$ 
& 84.8$^{\color{black}+0.0}$ 
& 82.7$^{\color{color+}+1.9}$ \\

\midrule
\textit{Zh}
& 84.0$^{\color{color-}-1.6}$ 
& 55.2$^{\color{color+}+2.4}$ 
& 87.2$^{\color{color+}+0.8}$ 
& 93.6$^{\color{color+}+0.8}$ 
& 83.2$^{\color{color+}+0.8}$ 
& 85.6$^{\color{color+}+0.8}$ 
& 81.5$^{\color{color+}+0.7}$ \\

\bottomrule
\end{tabular}
\label{tab:steering_results}
\end{table*}

Taken together, our analyses indicate that multilingual behavior in MoE models is governed by structured routing and expert utilization patterns.
Languages within the same family tend to share routing behavior, while expert usage exhibits a clear layer-wise stratification: early and late layers support language-specific understanding and generation, whereas middle layers primarily operate in a language-agnostic manner and mediate cross-lingual capacity sharing.
These findings suggest that multilingual capability in MoE models is largely governed by how effectively a language can access and leverage shared capacity.
Languages that align more closely with dominant languages benefit from stronger cross-lingual capacity transfer.
This structured view of routing and expert specialization naturally motivates an inference-time intervention that steers the routing behavior of target languages toward dominant languages within the same language family, while confining such alignment to the middle layers to preserve language-specific understanding and generation.

\subsection{Routing-Guided Steering}
\label{subsec:steering_method}
We implement steering by adaptively guiding the router toward experts
that are strongly associated with dominant languages,
and apply this adjustment exclusively to the middle layers.

Let $\mathcal{L}_{\mathrm{dom}} \subset \mathcal{L}$ denote the set of dominant languages.
Recall that $W_{l,e}^{(\ell)}$ denotes the normalized routing frequency of expert $e$ for language $\ell$ at layer $l$, which measures the strength of association between this expert and the corresponding language in the routing space.
We focus specifically on the language-shared experts of dominant languages, identified by
\begin{equation}
\label{eq:identify_shared}
\mathcal{E}_{l,\mathrm{shared}}
=
\left\{
i \;\middle|\;
\max_{\ell \in \mathcal{L}_{\mathrm{dom}}} W_{l,i}^{(\ell)} \le \theta
\right\}.
\end{equation}

During inference, we apply an expert-adaptive steering to the router logits of these experts in the middle layers $l \in \mathcal{L}_{\mathrm{mid}}$.
Specifically, the adjustment applied to each expert is modulated by its language association strength $W_{l,i}^{(\ell)}$ and by the magnitude of its original routing logit, such that experts that are more strongly aligned with the steering language and more relevant to the current input receive a larger bias:
\begin{equation}
\label{eq:steer}
\tilde{g}_{l,i}(x)
=
g_{l,i}(x)
+
\lambda \cdot W_{l,i}^{(\ell)} \cdot \lvert g_{l,i}(x) \rvert ,
\end{equation}
where $\lambda$ controls the steering strength,
and $\ell \in \mathcal{L}_{\mathrm{dom}}$ denotes the specific dominant language
serving as the steering source.
For early and late layers, routing logits remain unchanged.
Detailed experimental settings for the routing-guided steering are provided in
Appendix~\ref{app:routing-guided steering settings}.

\subsection{Steering Results}
\label{subsec:steering_results}
We evaluate our proposed method on the subset of PolyMath~\cite{PolyMath}, a representative math reasoning benchmark.
Table~\ref{tab:steering_results} reports results using steering signals
derived from two dominant languages, English and Chinese.
Overall, routing-guided steering yields consistent performance improvements across both high-resource and low-resource target languages.
Specifically, using English as the steering source leads to an average improvement of 1.9\%, while steering from Chinese achieves a more modest but consistent average gain of 0.7\%.
These results indicate that enhancing alignment with dominant languages can effectively improve cross-lingual capacity transfer, with the magnitude of improvement influenced by the choice of steering source.
More importantly, steering effectiveness strongly depends on linguistic proximity.
When English is used as the steering source, substantial gains are observed for languages sharing the Latin script, including French, Spanish, and Swahili.
In contrast, steering from English provides no effects for linguistically distant languages such as Japanese.
Conversely, using Chinese as the steering source successfully yields the improvements for Japanese, while providing smaller gains for languages from other families compared to English-based steering.
We additionally perform an ablation study by applying routing-guided steering to early and late layers.
Steering at these layers consistently degrades performance, as it disrupts language-specific capabilities essential for input understanding and output generation.
Detailed ablation results can be found in Appendix~\ref{app:addition-result-steering}.

\section{Conclusion}
\label{sec:conclusion}

In this work, we investigate multilingual processing in Mixture-of-Experts large language models
through a systematic analysis of routing behavior, expert specialization, and layerwise functional roles.
Our analysis reveals that multilingual capability in MoE models emerges from structured patterns of expert specialization and sharing.
Specifically, routing analysis reveals that expert selection patterns align with language family structure, while expert-level analysis exposes systematic differences in expert utilization across languages with different resource levels.
Moreover, these analyses uncover a clear layerwise organization, with language-specific processing concentrated in early and late layers.
Supported by layerwise intervention experiments, these findings motivate a routing-guided steering approach that aligns target languages toward dominant languages within the same language family through middle-layer experts, leading to consistent multilingual performance improvements.
\footnote{A discussion of related work is provided in Appendix~\ref{sec:related work}.}

\section*{Limitations}
\label{sec:limitation}

While this work provides a systematic analysis of how MoE models handle multilingual inputs, it is subject to several practical limitations.
First, our study is conducted on a representative and advanced large-scale MoE model.
As larger-scale MoE models continue to become more accessible, we plan to extend our analysis to substantially larger systems in order to further examine the generality and scaling behavior of our findings.
Second, although our results demonstrate that expert-level steering is a powerful mechanism for improving multilingual performance, our investigation is restricted to inference-time interventions.
Extending these insights to training-time or post-training optimization, where expert behaviors can be explicitly shaped and reinforced, constitutes a promising and complementary direction for future work.

\bibliography{reference}

@article{multilingual-moe,
  author       = {Lucas Bandarkar and
                  Chenyuan Yang and
                  Mohsen Fayyaz and
                  Junlin Hu and
                  Nanyun (Violet) Peng},
  title        = {Multilingual Routing in Mixture-of-Experts},
  journal      = {CoRR},
  volume       = {abs/2510.04694},
  year         = {2025},
  bibsource    = {dblp computer science bibliography, https://dblp.org}
}

@inproceedings{moe-2017,
  author       = {Noam Shazeer and
                  Azalia Mirhoseini and
                  Krzysztof Maziarz and
                  Andy Davis and
                  Quoc V. Le and
                  Geoffrey E. Hinton and
                  Jeff Dean},
  title        = {Outrageously Large Neural Networks: The Sparsely-Gated Mixture-of-Experts
                  Layer},
  booktitle    = {5th International Conference on Learning Representations, {ICLR} 2017,
                  Toulon, France, April 24-26, 2017, Conference Track Proceedings},
  publisher    = {OpenReview.net},
  year         = {2017},
  bibsource    = {dblp computer science bibliography, https://dblp.org}
}

@article{moe-survey,
  author       = {Weilin Cai and
                  Juyong Jiang and
                  Fan Wang and
                  Jing Tang and
                  Sunghun Kim and
                  Jiayi Huang},
  title        = {A Survey on Mixture of Experts in Large Language Models},
  journal      = {{IEEE} Trans. Knowl. Data Eng.},
  volume       = {37},
  number       = {7},
  pages        = {3896--3915},
  year         = {2025},
  bibsource    = {dblp computer science bibliography, https://dblp.org}
}

@article{moe-1991,
  author       = {Robert A. Jacobs and
                  Michael I. Jordan and
                  Steven J. Nowlan and
                  Geoffrey E. Hinton},
  title        = {Adaptive Mixtures of Local Experts},
  journal      = {Neural Comput.},
  volume       = {3},
  number       = {1},
  pages        = {79--87},
  year         = {1991},
  bibsource    = {dblp computer science bibliography, https://dblp.org}
}

@article{gpt-oss,
  author       = {OpenAI},
  title        = {gpt-oss-120b {\&} gpt-oss-20b Model Card},
  journal      = {CoRR},
  volume       = {abs/2508.10925},
  year         = {2025},
  bibsource    = {dblp computer science bibliography, https://dblp.org}
}

@article{Deepseek-R1,
  author       = {DeepSeek{-}AI},
  title        = {DeepSeek-R1: Incentivizing Reasoning Capability in LLMs via Reinforcement
                  Learning},
  journal      = {CoRR},
  volume       = {abs/2501.12948},
  year         = {2025}
}

@article{Babel,
  author       = {Yiran Zhao and
                  Chaoqun Liu and
                  Yue Deng and
                  Jiahao Ying and
                  Mahani Aljunied and
                  Zhaodonghui Li and
                  Lidong Bing and
                  Hou Pong Chan and
                  Yu Rong and
                  Deli Zhao and
                  Wenxuan Zhang},
  title        = {Babel: Open Multilingual Large Language Models Serving Over 90{\%}
                  of Global Speakers},
  journal      = {CoRR},
  volume       = {abs/2503.00865},
  year         = {2025}
}

@article{Qwen3,
  author       = {An Yang and
                  Anfeng Li and
                  Baosong Yang and
                  Beichen Zhang and
                  Binyuan Hui and
                  Bo Zheng and
                  Bowen Yu and
                  Chang Gao and
                  Chengen Huang and
                  Chenxu Lv and
                  Chujie Zheng and
                  Dayiheng Liu and
                  Fan Zhou and
                  Fei Huang and
                  Feng Hu and
                  Hao Ge and
                  Haoran Wei and
                  Huan Lin and
                  Jialong Tang and
                  Jian Yang and
                  Jianhong Tu and
                  Jianwei Zhang and
                  Jian Yang and
                  Jiaxi Yang and
                  Jingren Zhou and
                  Junyang Lin and
                  Kai Dang and
                  Keqin Bao and
                  Kexin Yang and
                  Le Yu and
                  Lianghao Deng and
                  Mei Li and
                  Mingfeng Xue and
                  Mingze Li and
                  Pei Zhang and
                  Peng Wang and
                  Qin Zhu and
                  Rui Men and
                  Ruize Gao and
                  Shixuan Liu and
                  Shuang Luo and
                  Tianhao Li and
                  Tianyi Tang and
                  Wenbiao Yin and
                  Xingzhang Ren and
                  Xinyu Wang and
                  Xinyu Zhang and
                  Xuancheng Ren and
                  Yang Fan and
                  Yang Su and
                  Yichang Zhang and
                  Yinger Zhang and
                  Yu Wan and
                  Yuqiong Liu and
                  Zekun Wang and
                  Zeyu Cui and
                  Zhenru Zhang and
                  Zhipeng Zhou and
                  Zihan Qiu},
  title        = {Qwen3 Technical Report},
  journal      = {CoRR},
  volume       = {abs/2505.09388},
  year         = {2025}
}

@article{deepseek-v3,
  author       = {DeepSeek{-}AI},
  title        = {DeepSeek-V3 Technical Report},
  journal      = {CoRR},
  volume       = {abs/2412.19437},
  year         = {2024}
}

@inproceedings{deepseekmoe,
  author       = {Damai Dai and
                  Chengqi Deng and
                  Chenggang Zhao and
                  R. X. Xu and
                  Huazuo Gao and
                  Deli Chen and
                  Jiashi Li and
                  Wangding Zeng and
                  Xingkai Yu and
                  Y. Wu and
                  Zhenda Xie and
                  Y. K. Li and
                  Panpan Huang and
                  Fuli Luo and
                  Chong Ruan and
                  Zhifang Sui and
                  Wenfeng Liang},
  title        = {DeepSeekMoE: Towards Ultimate Expert Specialization in Mixture-of-Experts
                  Language Models},
  booktitle    = {{ACL} {(1)}},
  pages        = {1280--1297},
  publisher    = {Association for Computational Linguistics},
  year         = {2024}
}

@article{activation_1,
  author       = {Lisa Schut and
                  Yarin Gal and
                  Sebastian Farquhar},
  title        = {Do Multilingual LLMs Think In English?},
  journal      = {CoRR},
  volume       = {abs/2502.15603},
  year         = {2025}
}

@inproceedings{activation_2,
  author       = {Chris Wendler and
                  Veniamin Veselovsky and
                  Giovanni Monea and
                  Robert West},
  title        = {Do Llamas Work in English? On the Latent Language of Multilingual
                  Transformers},
  booktitle    = {{ACL} {(1)}},
  pages        = {15366--15394},
  publisher    = {Association for Computational Linguistics},
  year         = {2024}
}

@article{AbstractThought,
  author       = {Yuxin Chen and
                  Yiran Zhao and
                  Yang Zhang and
                  An Zhang and
                  Kenji Kawaguchi and
                  Shafiq Joty and
                  Junnan Li and
                  Tat{-}Seng Chua and
                  Michael Qizhe Shieh and
                  Wenxuan Zhang},
  title        = {The Emergence of Abstract Thought in Large Language Models Beyond
                  Any Language},
  journal      = {CoRR},
  volume       = {abs/2506.09890},
  year         = {2025}
}

@inproceedings{yiran_howdo,
  author       = {Yiran Zhao and
                  Wenxuan Zhang and
                  Guizhen Chen and
                  Kenji Kawaguchi and
                  Lidong Bing},
  title        = {How do Large Language Models Handle Multilingualism?},
  booktitle    = {NeurIPS},
  year         = {2024}
}

@article{attention_1,
  author       = {Sai Gopinath and
                  Joselyn Rodriguez},
  title        = {Probing self-attention in self-supervised speech models for cross-linguistic
                  differences},
  journal      = {CoRR},
  volume       = {abs/2409.03115},
  year         = {2024}
}

@inproceedings{attention_2,
  author       = {Weicheng Ma and
                  Kai Zhang and
                  Renze Lou and
                  Lili Wang and
                  Soroush Vosoughi},
  title        = {Contributions of Transformer Attention Heads in Multi- and Cross-lingual
                  Tasks},
  booktitle    = {{ACL/IJCNLP} {(1)}},
  pages        = {1956--1966},
  publisher    = {Association for Computational Linguistics},
  year         = {2021}
}

@inproceedings{attention_3,
  author       = {Elena Voita and
                  David Talbot and
                  Fedor Moiseev and
                  Rico Sennrich and
                  Ivan Titov},
  title        = {Analyzing Multi-Head Self-Attention: Specialized Heads Do the Heavy
                  Lifting, the Rest Can Be Pruned},
  booktitle    = {{ACL} {(1)}},
  pages        = {5797--5808},
  publisher    = {Association for Computational Linguistics},
  year         = {2019}
}

@inproceedings{MGSM,
  author       = {Freda Shi and
                  Mirac Suzgun and
                  Markus Freitag and
                  Xuezhi Wang and
                  Suraj Srivats and
                  Soroush Vosoughi and
                  Hyung Won Chung and
                  Yi Tay and
                  Sebastian Ruder and
                  Denny Zhou and
                  Dipanjan Das and
                  Jason Wei},
  title        = {Language models are multilingual chain-of-thought reasoners},
  booktitle    = {{ICLR}},
  publisher    = {OpenReview.net},
  year         = {2023}
}

@inproceedings{Belebele,
  author       = {Lucas Bandarkar and
                  Davis Liang and
                  Benjamin Muller and
                  Mikel Artetxe and
                  Satya Narayan Shukla and
                  Donald Husa and
                  Naman Goyal and
                  Abhinandan Krishnan and
                  Luke Zettlemoyer and
                  Madian Khabsa},
  title        = {The Belebele Benchmark: a Parallel Reading Comprehension Dataset in
                  122 Language Variants},
  booktitle    = {{ACL} {(1)}},
  pages        = {749--775},
  publisher    = {Association for Computational Linguistics},
  year         = {2024}
}

@article{PolyMath,
  author       = {Yiming Wang and
                  Pei Zhang and
                  Jialong Tang and
                  Haoran Wei and
                  Baosong Yang and
                  Rui Wang and
                  Chenshu Sun and
                  Feitong Sun and
                  Jiran Zhang and
                  Junxuan Wu and
                  Qiqian Cang and
                  Yichang Zhang and
                  Fei Huang and
                  Junyang Lin and
                  Fei Huang and
                  Jingren Zhou},
  title        = {PolyMath: Evaluating Mathematical Reasoning in Multilingual Contexts},
  journal      = {CoRR},
  volume       = {abs/2504.18428},
  year         = {2025}
}

@article{flore200,
  author       = {Marta R. Costa{-}juss{\`{a}} and
                  James Cross and
                  Onur {\c{C}}elebi and
                  Maha Elbayad and
                  Kenneth Heafield and
                  Kevin Heffernan and
                  Elahe Kalbassi and
                  Janice Lam and
                  Daniel Licht and
                  Jean Maillard and
                  Anna Y. Sun and
                  Skyler Wang and
                  Guillaume Wenzek and
                  Al Youngblood and
                  Bapi Akula and
                  Lo{\"{\i}}c Barrault and
                  Gabriel Mejia Gonzalez and
                  Prangthip Hansanti and
                  John Hoffman and
                  Semarley Jarrett and
                  Kaushik Ram Sadagopan and
                  Dirk Rowe and
                  Shannon Spruit and
                  Chau Tran and
                  Pierre Andrews and
                  Necip Fazil Ayan and
                  Shruti Bhosale and
                  Sergey Edunov and
                  Angela Fan and
                  Cynthia Gao and
                  Vedanuj Goswami and
                  Francisco Guzm{\'{a}}n and
                  Philipp Koehn and
                  Alexandre Mourachko and
                  Christophe Ropers and
                  Safiyyah Saleem and
                  Holger Schwenk and
                  Jeff Wang},
  title        = {No Language Left Behind: Scaling Human-Centered Machine Translation},
  journal      = {CoRR},
  volume       = {abs/2207.04672},
  year         = {2022}
}

@inproceedings{xquad,
  author       = {Mikel Artetxe and
                  Sebastian Ruder and
                  Dani Yogatama},
  title        = {On the Cross-lingual Transferability of Monolingual Representations},
  booktitle    = {{ACL}},
  pages        = {4623--4637},
  publisher    = {Association for Computational Linguistics},
  year         = {2020}
}

@article{zhang2024seallms,
  title={Seallms 3: Open foundation and chat multilingual large language models for southeast asian languages},
  author={Zhang, Wenxuan and Chan, Hou Pong and Zhao, Yiran and Aljunied, Mahani and Wang, Jianyu and Liu, Chaoqun and Deng, Yue and Hu, Zhiqiang and Xu, Weiwen and Chia, Yew Ken and others},
  journal={arXiv preprint arXiv:2407.19672},
  year={2024}
}

@article{dou2025sailor2,
  title={Sailor2: Sailing in South-East Asia with Inclusive Multilingual LLMs},
  author={Dou, Longxu and Liu, Qian and Zhou, Fan and Chen, Changyu and Wang, Zili and Jin, Ziqi and Liu, Zichen and Zhu, Tongyao and Du, Cunxiao and Yang, Penghui and others},
  journal={arXiv preprint arXiv:2502.12982},
  year={2025}
}

@article{zhou2025disparities,
  title={Disparities in LLM Reasoning Accuracy and Explanations: A Case Study on African American English},
  author={Zhou, Runtao and Wan, Guangya and Gabriel, Saadia and Li, Sheng and Gates, Alexander J and Sap, Maarten and Hartvigsen, Thomas},
  journal={arXiv preprint arXiv:2503.04099},
  year={2025}
}

@inproceedings{zhu2024multilingual,
  title={Multilingual Machine Translation with Large Language Models: Empirical Results and Analysis},
  author={Zhu, Wenhao and Liu, Hongyi and Dong, Qingxiu and Xu, Jingjing and Huang, Shujian and Kong, Lingpeng and Chen, Jiajun and Li, Lei},
  booktitle={Findings of the Association for Computational Linguistics: NAACL 2024},
  pages={2765--2781},
  year={2024}
}

@inproceedings{she2024mapo,
  title={MAPO: Advancing Multilingual Reasoning through Multilingual-Alignment-as-Preference Optimization},
  author={She, Shuaijie and Zou, Wei and Huang, Shujian and Zhu, Wenhao and Liu, Xiang and Geng, Xiang and Chen, Jiajun},
  booktitle={Proceedings of the 62nd Annual Meeting of the Association for Computational Linguistics (Volume 1: Long Papers)},
  pages={10015--10027},
  year={2024}
}

@article{shimabucoro2025post,
  title={A Post-trainer's Guide to Multilingual Training Data: Uncovering Cross-lingual Transfer Dynamics},
  author={Shimabucoro, Lu{\'\i}sa and Ustun, Ahmet and Fadaee, Marzieh and Ruder, Sebastian},
  journal={arXiv preprint arXiv:2504.16677},
  year={2025}
}

@article{ko2025understand,
  title={Understand, Solve and Translate: Bridging the Multilingual Mathematical Reasoning Gap},
  author={Ko, Hyunwoo and Son, Guijin and Choi, Dasol},
  journal={arXiv preprint arXiv:2501.02448},
  year={2025}
}

@article{zhao2024lens,
  title={Lens: Rethinking Multilingual Enhancement for Large Language Models},
  author={Zhao, Weixiang and Hu, Yulin and Guo, Jiahe and Sui, Xingyu and Wu, Tongtong and Deng, Yang and Zhao, Yanyan and Qin, Bing and Che, Wanxiang and Liu, Ting},
  journal={arXiv preprint arXiv:2410.04407},
  year={2024}
}

@inproceedings{zhang2024enhancing,
  author       = {Yuanchi Zhang and
                  Yile Wang and
                  Zijun Liu and
                  Shuo Wang and
                  Xiaolong Wang and
                  Peng Li and
                  Maosong Sun and
                  Yang Liu},
  editor       = {Lun{-}Wei Ku and
                  Andre Martins and
                  Vivek Srikumar},
  title        = {Enhancing Multilingual Capabilities of Large Language Models through
                  Self-Distillation from Resource-Rich Languages},
  booktitle    = {Proceedings of the 62nd Annual Meeting of the Association for Computational
                  Linguistics (Volume 1: Long Papers), {ACL} 2024, Bangkok, Thailand,
                  August 11-16, 2024},
  pages        = {11189--11204},
  publisher    = {Association for Computational Linguistics},
  year         = {2024},
  bibsource    = {dblp computer science bibliography, https://dblp.org}
}

@inproceedings{huo2025enhancing,
  title={Enhancing Non-English Capabilities of English-Centric Large Language Models Through Deep Supervision Fine-Tuning},
  author={Huo, Wenshuai and Feng, Xiaocheng and Huang, Yichong and Fu, Chengpeng and Li, Baohang and Ye, Yangfan and Zhang, Zhirui and Tu, Dandan and Tang, Duyu and Lu, Yunfei and others},
  booktitle={Proceedings of the AAAI Conference on Artificial Intelligence},
  volume={39},
  number={23},
  pages={24185--24193},
  year={2025}
}

@inproceedings{ruan2025layalign,
  title={LayAlign: Enhancing Multilingual Reasoning in Large Language Models via Layer-Wise Adaptive Fusion and Alignment Strategy},
  author={Ruan, Zhiwen and Li, Yixia and Zhu, He and Wang, Longyue and Luo, Weihua and Zhang, Kaifu and Chen, Yun and Chen, Guanhua},
  booktitle={Findings of the Association for Computational Linguistics: NAACL 2025},
  pages={1481--1495},
  year={2025}
}

@inproceedings{fan2025slam,
  title={SLAM: Towards Efficient Multilingual Reasoning via Selective Language Alignment},
  author={Fan, Yuchun and Mu, Yongyu and Wang, Yilin and Huang, Lei and Ruan, Junhao and Li, Bei and Xiao, Tong and Huang, Shujian and Feng, Xiaocheng and Zhu, Jingbo},
  booktitle={Proceedings of the 31st International Conference on Computational Linguistics},
  pages={9499--9515},
  year={2025}
}

@inproceedings{qin2023cross,
  title={Cross-lingual Prompting: Improving Zero-shot Chain-of-Thought Reasoning across Languages},
  author={Qin, Libo and Chen, Qiguang and Wei, Fuxuan and Huang, Shijue and Che, Wanxiang},
  booktitle={Proceedings of the 2023 Conference on Empirical Methods in Natural Language Processing},
  pages={2695--2709},
  year={2023}
}

@inproceedings{zhang2024autocap,
  title={AutoCAP: Towards Automatic Cross-lingual Alignment Planning for Zero-shot Chain-of-Thought},
  author={Zhang, Yongheng and Chen, Qiguang and Li, Min and Che, Wanxiang and Qin, Libo},
  booktitle={Findings of the Association for Computational Linguistics ACL 2024},
  pages={9191--9200},
  year={2024}
}

@article{yong2025crosslingual,
  title={Crosslingual Reasoning through Test-Time Scaling},
  author={Yong, Zheng-Xin and Adilazuarda, M Farid and Mansurov, Jonibek and Zhang, Ruochen and Muennighoff, Niklas and Eickhoff, Carsten and Winata, Genta Indra and Kreutzer, Julia and Bach, Stephen H and Aji, Alham Fikri},
  journal={arXiv preprint arXiv:2505.05408},
  year={2025}
}

@article{gao2025could,
  title={Could Thinking Multilingually Empower LLM Reasoning?},
  author={Gao, Changjiang and Huang, Xu and Zhu, Wenhao and Huang, Shujian and Li, Lei and Yuan, Fei},
  journal={arXiv preprint arXiv:2504.11833},
  year={2025}
}

@article{tran2025scaling,
  title={Scaling Test-time Compute for Low-resource Languages: Multilingual Reasoning in LLMs},
  author={Tran, Khanh-Tung and O'Sullivan, Barry and Nguyen, Hoang D},
  journal={arXiv preprint arXiv:2504.02890},
  year={2025}
}

@article{yu2025cross,
  title={Cross-Lingual Consistency: A Novel Inference Framework for Advancing Reasoning in Large Language Models},
  author={Yu, Zhiwei and Li, Tuo and Wang, Changhong and Chen, Hui and Zhou, Lang},
  journal={arXiv preprint arXiv:2504.01857},
  year={2025}
}

@article{son2025linguistic,
  title={Linguistic generalizability of test-time scaling in mathematical reasoning},
  author={Son, Guijin and Hong, Jiwoo and Ko, Hyunwoo and Thorne, James},
  journal={arXiv preprint arXiv:2502.17407},
  year={2025}
}

@inproceedings{tang2024language,
  title     = {Language-Specific Neurons: The Key to Multilingual Capabilities in Large Language Models},
  author    = {Tang, Tianyi and Luo, Wenyang and Huang, Haoyang and Zhang, Dongdong and Wang, Xiaolei and Zhao, Wayne Xin and Wei, Furu and Wen, Ji-Rong},
  booktitle = {Proceedings of the 62nd Annual Meeting of the Association for Computational Linguistics (Volume 1: Long Papers)},
  pages     = {5701--5715},
  year      = {2024}
}

@inproceedings{kojima2024multilingual,
  title     = {On the Multilingual Ability of Decoder-based Pre-trained Language Models: Finding and Controlling Language-Specific Neurons},
  author    = {Kojima, Takeshi and Okimura, Itsuki and Iwasawa, Yusuke and Yanaka, Hitomi and Matsuo, Yutaka},
  booktitle = {Proceedings of the 2024 Conference of the North American Chapter of the Association for Computational Linguistics: Human Language Technologies (Volume 1: Long Papers)},
  pages     = {6919--6971},
  year      = {2024},
  publisher = {Association for Computational Linguistics}
}

@article{saito2024we,
  title   = {Why We Build Local Large Language Models: An Observational Analysis from 35 Japanese and Multilingual LLMs},
  author  = {Saito, Koshiro and Mizuki, Sakae and Ohi, Masanari and Nakamura, Taishi and Shiotani, Taihei and Maeda, Koki and Ma, Youmi and Hattori, Kakeru and Fujii, Kazuki and Okamoto, Takumi and others},
  journal = {arXiv preprint arXiv:2412.14471},
  year    = {2024}
}

@inproceedings{zhang2024unveiling,
  title     = {Unveiling Linguistic Regions in Large Language Models},
  author    = {Zhang, Zhihao and Zhao, Jun and Zhang, Qi and Gui, Tao and Huang, Xuan-Jing},
  booktitle = {Proceedings of the 62nd Annual Meeting of the Association for Computational Linguistics (Volume 1: Long Papers)},
  pages     = {6228--6247},
  year      = {2024}
}

@article{wu2024semantic,
  title   = {The Semantic Hub Hypothesis: Language Models Share Semantic Representations Across Languages and Modalities},
  author  = {Wu, Zhaofeng and Yu, Xinyan Velocity and Yogatama, Dani and Lu, Jiasen and Kim, Yoon},
  journal = {arXiv preprint arXiv:2411.04986},
  year    = {2024}
}

@article{wang2024sharing,
  title   = {Sharing Matters: Analysing Neurons Across Languages and Tasks in LLMs},
  author  = {Wang, Weixuan and Haddow, Barry and Wu, Minghao and Peng, Wei and Birch, Alexandra},
  journal = {arXiv preprint arXiv:2406.09265},
  year    = {2024}
}

@article{brinkmann2025large,
  title   = {Large Language Models Share Representations of Latent Grammatical Concepts Across Typologically Diverse Languages},
  author  = {Brinkmann, Jannik and Wendler, Chris and Bartelt, Christian and Mueller, Aaron},
  journal = {arXiv preprint arXiv:2501.06346},
  year    = {2025}
}

\clearpage
\onecolumn

\renewcommand{\cftsecfont}{\normalsize} 
\renewcommand{\cftsubsecfont}{\normalsize} 
\renewcommand{\cftbeforesecskip}{12pt}      
\renewcommand{\cftbeforesubsecskip}{12pt}  

\renewcommand{\contentsname}{Appendix}
\addtocontents{toc}{\protect\setcounter{tocdepth}{3}}
\appendix

\clearpage
\twocolumn

\appendix

\section{Experimental Settings}
\label{app:experimental settings}
\subsection{Language Family-Structured Grouping}
\label{app:language family settings}
\paragraph{Language Selection.}
We select a diverse set of 13 languages from \textsc{Belebele} benchmark to analyze routing behaviors across linguistic families and scripts. Specifically, we included Indo-European languages (English, Spanish, French, German, Portuguese, Italian) to observe intra-family dynamics; Slavic (Russian, Ukrainian) and Perso-Arabic (Arabic, Persian) groups to examine alignment based on script and lexical similarity; and CJK-related languages (Simplified/Traditional Chinese, Japanese) to investigate strategies for logographic systems.
 
\paragraph{Experiment Settings.}
For each target language $s$, we utilize the complete test split of the \textsc{Belebele} dataset, comprising 900 parallel reading comprehension samples. To capture routing behaviors, we perform a forward pass using the Qwen3-30B-A3B model and record the expert selection frequencies. These selections are aggregated to construct the routing distribution in Eq.~(\ref{eq:routing_distribution}). Finally, to quantify the alignment between language pairs, we compute the cross-language routing similarity defined in Eq.~(\ref{eq:routing_similarity}) at the final MoE layer.

\subsection{Expert Specialization}
\label{app:expert specialization settings}

\paragraph{Language Selection.}
We focused our analysis on 10 representative languages: Arabic, Bengali, German, English, Spanish, French, Japanese, Korean, Swahili, and Chinese, as detailed in Section~\ref{sec:analysis_setup}.

\paragraph{Experiment Settings.}
For each language, we utilized the complete test split of the \textsc{Belebele} dataset. By performing a forward pass with the Qwen3-30B-A3B model, we calculated the expert selection frequencies. We first identified the top $K=15$ language-related experts for each language (Eq.~\ref{eq:language-related-experts}), and subsequently classified an expert as exclusive if its normalized affinity exceeded a concentration threshold of $\theta = 0.4$ (Eq.~\ref{eq:language-exclusive-experts}). This threshold was selected to rigorously isolate experts driven by specific linguistic signals while filtering out those broadly shared across languages.

\subsection{Expert Intervention}
\label{app:expert intervention settings}

To ensure a fair comparison of functional roles across different network depths, we standardize the intervention scope to a fixed window of five layers. We designate indices 0 to 4 as early layers, indices 22 to 26 as middle layers, and indices 43 to 47 as late layers. Furthermore, the selection of exclusive experts in this analysis follows the same criteria as detailed in Appendix~\ref{app:expert specialization settings}.

\subsection{Routing-Guided Steering}
\label{app:routing-guided steering settings}

\paragraph{Steering Parameter Construction.}
The core of our approach relies on the computation of their normalized routing frequencies $W_{l,i}^{(\ell)}$ and the precise identification of language-shared experts. Instead of a fixed steering coefficient, we utilize these fine-grained expert routing frequencies to guide the routing. To compute these statistics robustly, we utilize the \textsc{MGSM} dataset as the source of routing data. Specifically, we perform a forward pass on the MGSM samples corresponding to the dominant languages ($\mathcal{L}_{\text{dom}} = \{\texttt{en}, \texttt{zh}\}$), recording the expert selection frequencies. Based on these statistics, we identify the language-related experts by selecting the Top-$K$ candidates with the highest frequencies for each language, setting $K=20$ in Eq.~(\ref{eq:language-related-experts}). 
Subsequently, we calculate the normalized routing frequencies $W_{l,i}^{(\ell)}$ for these experts and identify the language-shared experts using a concentration threshold of $\theta=0.7$ in Eq.~(\ref{eq:shared_experts}). Experts whose maximum normalized routing frequencies across languages does not exceed this threshold are classified as shared experts, serving as the target experts for our steering intervention.

\paragraph{Inference and Evaluation.}
During inference, we apply the steering intervention according to Eq.~(\ref{eq:steer}) by modifying the router logits of the identified shared experts. This involves injecting a adaptive bias term $\lambda \cdot W_{l,e}^{(s)} \cdot \lvert g_{l,e}(x) \rvert$, which effectively encourages the router to utilize shared capacity relevant to the dominant languages. As established in our layerwise analysis (Section~\ref{subsec:routing_findings} \& \ref{subsec:expert_findings}) and intervention analysis (Section~\ref{sec:intervention}), this intervention is strictly applied to the middle layers from 10 to 39 to foster general reasoning transfer without disrupting the language-specific understanding and generation mechanisms located in the early and late layers, respectively. We evaluate the efficacy of this method on the low subset of \textsc{PolyMath} benchmark and report the results under the optimal configuration of the $\lambda$ coefficient. The hyperparameter study of $\lambda$ is detailed in Section~\ref{subsec:hyper}.
\section{Supplementary Metrics}
\label{app:supplementary metrics}
\subsection{Routing Entropy}
\label{app:routing-entropy}

As a complementary analysis, we also measure routing dispersion using routing entropy, 
which summarizes how broadly routing mass is distributed over experts for a given language.

For a language $s \in \mathcal{L}$ and layer $l$, 
let $\mathcal{D}_s$ denote the set of tokens in language $s$, with $N_s = |\mathcal{D}_s|$.
We define the average routing distribution over experts as
\begin{equation}
p_{l,e}^{(s)} = \frac{1}{K\,N_s} \sum_{x \in \mathcal{D}_s} 
\mathbb{I}\!\left(e \in \mathcal{S}_l(x)\right),
\label{eq:routing-dist-app}
\end{equation}
where $K$ is the number of selected experts under Top-$K$ routing.

Routing entropy at layer $l$ is then computed as
\begin{equation}
H_l^{(s)} = - \sum_{e \in \mathcal{E}} p_{l,e}^{(s)} \log \big(p_{l,e}^{(s)} \big).
\label{eq:routing-entropy-app}
\end{equation}

To obtain a layer-agnostic measure, we further average routing entropy across layers:
\begin{equation}
\bar{H}^{(s)} = \frac{1}{L}\sum_{l=0}^{L-1} H_l^{(s)}.
\label{eq:mean-routing-entropy-app}
\end{equation}

Higher routing entropy indicates that tokens from language $s$ are more evenly distributed across experts, 
whereas lower entropy suggests more concentrated expert usage.

\section{Additional Results}
\label{app:findings}

\subsection{Routing Entropy}
\label{app:routing-entropy-findings}

\begin{figure}[t]
  \centering
  \includegraphics[width=\columnwidth]{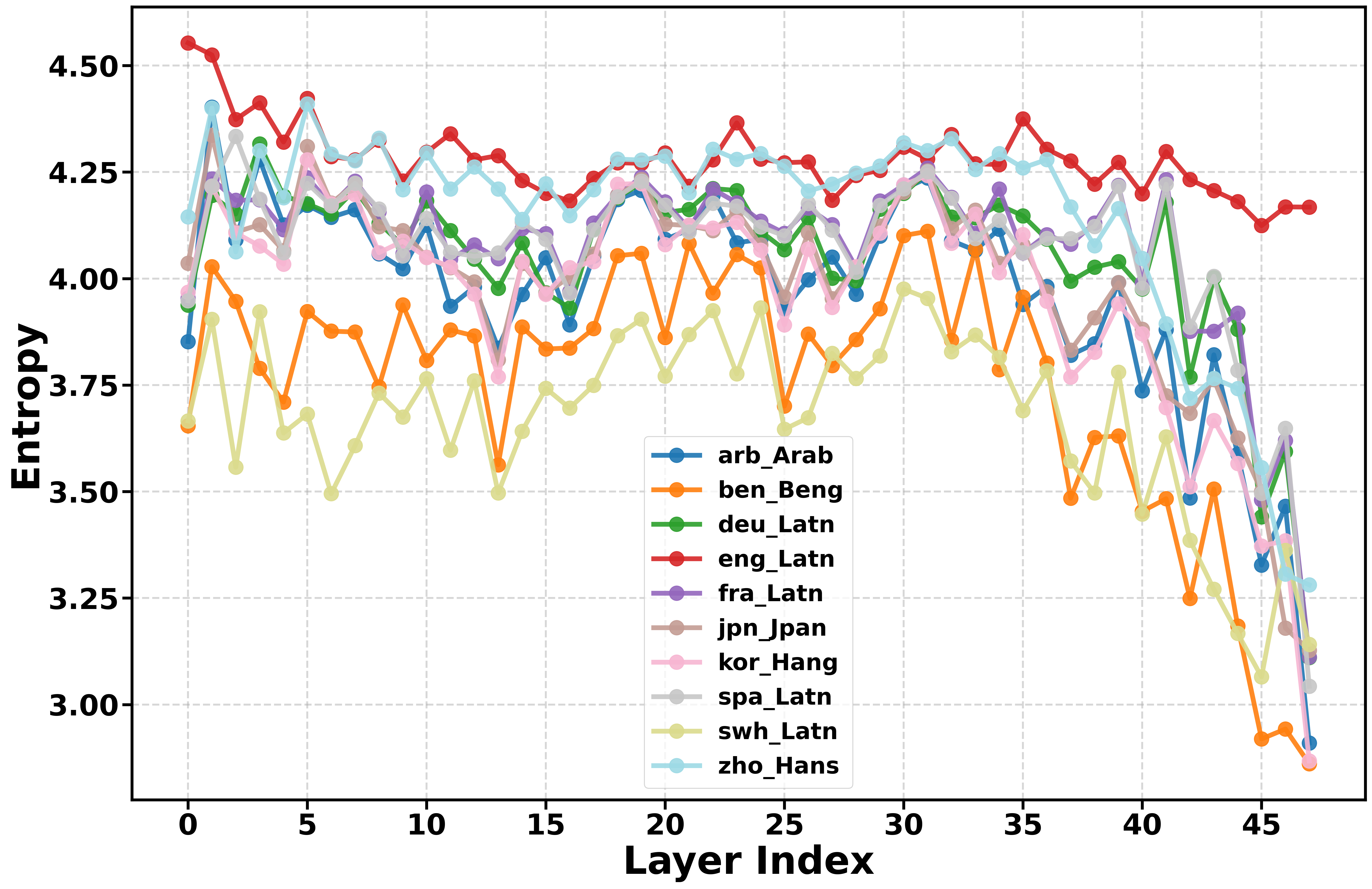}
  \caption{Routing entropy per layer for each language.
  \label{fig:entropy_curve}}
\end{figure}

We first highlight a key observation based on routing entropy:
languages are clearly separated into distinct groups with systematically different entropy levels.
Specifically, routing entropy consistently partitions languages into dominant, high-resource, and low-resource groups,
providing supporting evidence that MoE models are sensitive to language resource differences.

Across languages, dominant languages exhibit the highest routing entropy,
indicating that their tokens are distributed across a broad set of experts.
High-resource languages show moderately high entropy,
suggesting effective utilization of shared expert capacity.
In contrast, low-resource languages consistently exhibit substantially lower routing entropy,
reflecting more concentrated expert usage.
This clear stratification aligns with the language resource categories analyzed in the main text.

At the layer level, routing entropy remains relatively stable in the early and middle layers
and decreases toward the top of the network.
This trend is consistent with the layerwise expert specialization patterns reported in the main text,
where expert routing becomes increasingly concentrated in later layers.

Overall, routing entropy provides complementary evidence that MoE models allocate expert capacity unevenly across languages
in a manner that reflects language resource levels.
However, as routing entropy does not distinguish between language-exclusive and shared experts,
we treat it as a supplementary signal and rely primarily on expert specialization
and cross-language routing similarity for the main analysis of multilingual routing behavior.

\subsection{Expert Specialization}
\label{app:specialization}
In Section~\ref{sec:expert_analysis}, we discussed the distribution of language-exclusive experts averaged across language groups. Here, we provide a complementary fine-grained analysis by visualizing the expert specialization patterns for each individual language.

\begin{figure}[h]
    \centering
    \includegraphics[width=\linewidth]{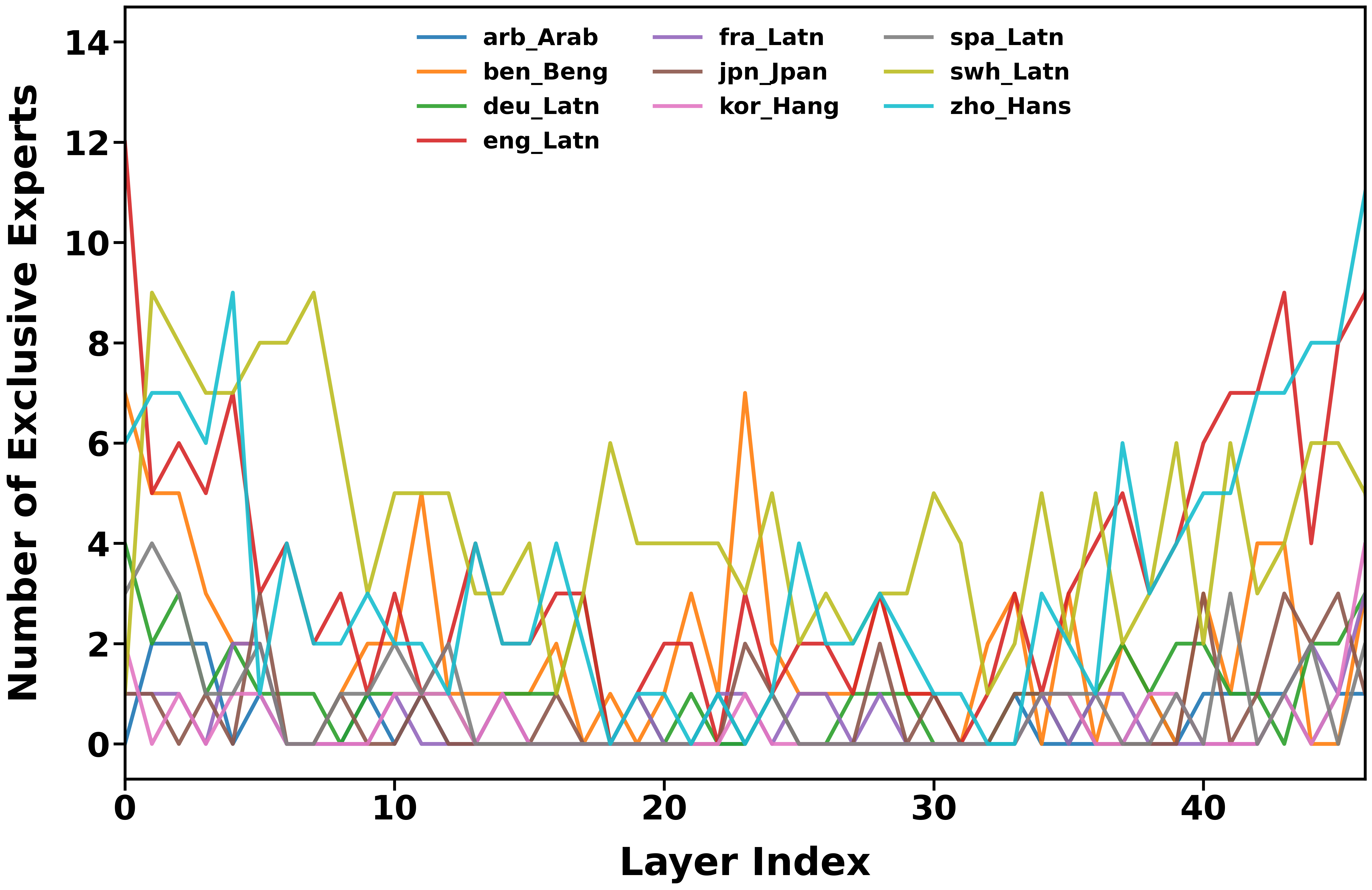}
    \caption{Layer-wise distribution of language-exclusive experts for each of the 10 investigated languages.}
    \label{fig:layer_exclusive_individual}
\end{figure}

\paragraph{Layer-wise Distribution.}
Figure~\ref{fig:layer_exclusive_individual} illustrates the number of exclusive experts per layer for all 10 languages individually. Consistent with the group-level findings in Section~\ref{subsec:expert_findings}, almost all languages exhibit a characteristic U-shaped pattern: expert specialization is prominent in the early and late layers while remaining minimal in the middle layers.

\begin{figure}[t!]
  \centering
  \includegraphics[width=\linewidth]{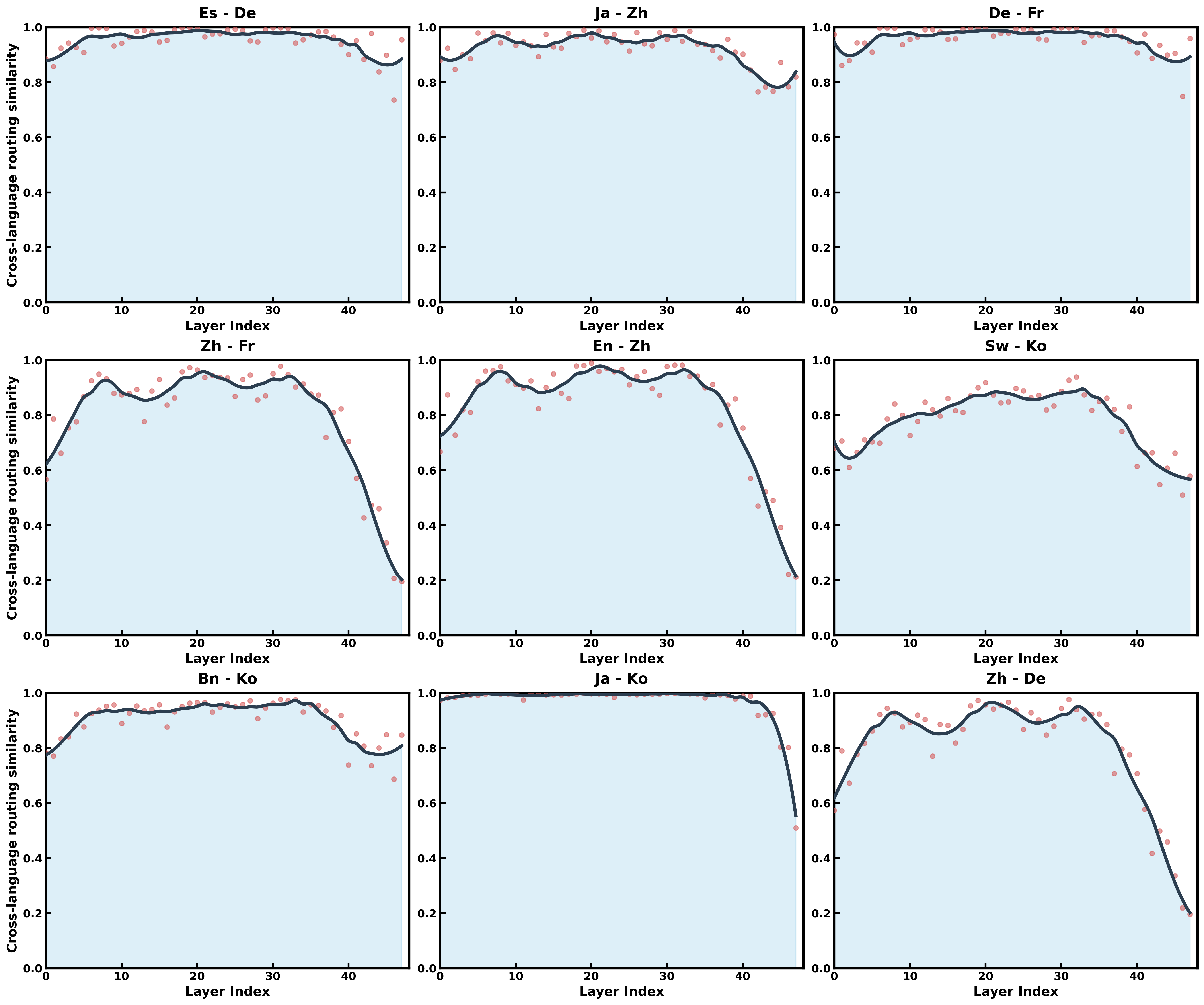}
  \caption{Layer-wise cross-language routing similarity for varying language pairs.}
  \label{fig:summary_3x3_grid}
\end{figure}

\paragraph{Layer-wise Routing Divergence.}
To further substantiate the functional stratification observed in expert counts, Figure~\ref{fig:summary_3x3_grid} presents the layer-wise cross-language routing similarity across nine representative language pairs. The first row displays pairs of linguistically related languages. These pairs exhibit a relatively flattened $\bigcap$-shape with lower overall divergence, confirming that related languages activate similar expert pathways. The second row highlights pairs from distant language families. Here, the $\bigcap$-shaped pattern is most pronounced, with significantly higher divergence in the early and late layers, reflecting distinct processing needs for unrelated languages. The third row shows randomly selected pairs. Regardless of linguistic proximity, the $\bigcap$-shaped trend remains robust across all combinations.
Collectively, these plots reinforce that intermediate layers are largely language-agnostic, whereas early and late layers drive language-specific processing.

\paragraph{Total Exclusive Experts.}
\begin{figure}[h]
    \centering
    \includegraphics[width=\linewidth]{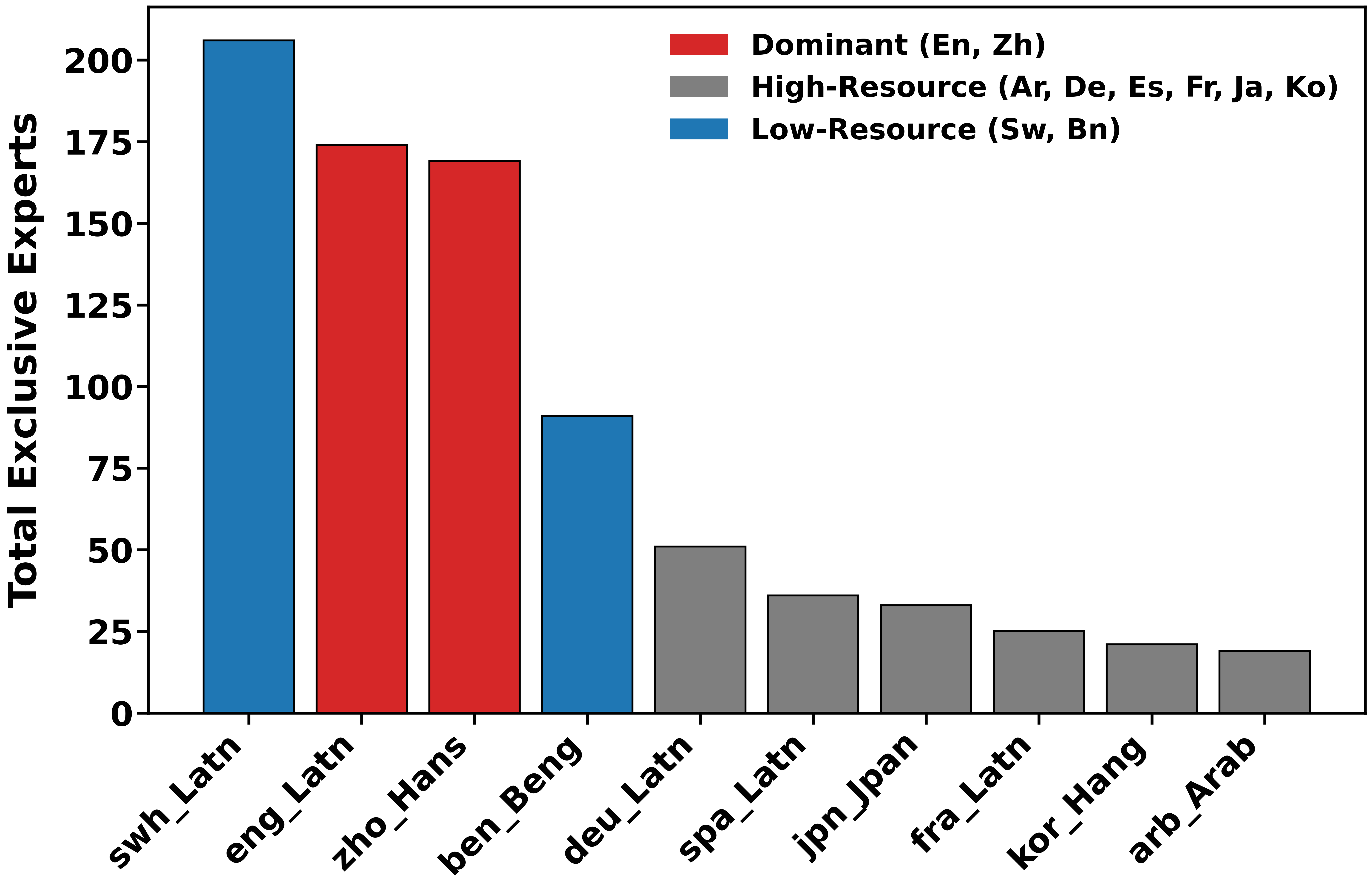}
    \caption{Total count of language-exclusive experts aggregated across all layers for each language, categorized by resource level.}
    \label{fig:total_exclusive}
\end{figure}
We further aggregate the number of exclusive experts across all layers to compare the overall degree of specialization, as shown in Figure~\ref{fig:total_exclusive}. The results highlight a stark disparity rooted in language resource levels. Low-resource languages possess the highest number of exclusive experts, suggesting they are less integrated into the shared expert space compared to high-resource languages, which effectively leverage shared capacity.

\subsection{Expert Intervention}
\label{app:addition-expert-intervention}

\begin{table*}[p!]
\centering
\renewcommand{\arraystretch}{1.3}
\setlength{\tabcolsep}{4.5pt}
\caption{Layer-wise intervention effects on accuracy in MGSM. We mask language-exclusive experts of target languages at early, middle, and late layers. Significant drops are highlighted in red.}
\small
\begin{tabular}{c|cccccccc}
\toprule
\multirow{2}{*}{\shortstack[c]{\textbf{Intervened} \\ \textbf{Layers}}} & \multicolumn{8}{c}{\textbf{Languages}} \\
 & {Bn} & {De} & {En} & {Es} & {Fr} & {Ja} & {Sw} & {Zh} \\
\midrule

\rowcolor{lightgray}
- & 84.8 & 88.8 & 92.8 & 92.8 & 86.4 & 85.6 & 54.8 & 90.0 \\

\midrule
\multicolumn{9}{c}{\textbf{\textit{Target Lang: Bengali (Bn)}}} \\
\midrule
\textit{Early} & \cellcolor{lightred}53.6$^{\color{color-}-31.2}$ & 88.4$^{\color{color-}-0.4}$ & 94.4$^{\color{color+}+1.6}$ & 94.0$^{\color{color+}+1.2}$ & 88.4$^{\color{color+}+2.0}$ & 84.8$^{\color{color-}-0.8}$ & 34.4$^{\color{color-}-20.4}$ & 88.0$^{\color{color-}-2.0}$ \\
\textit{Middle} & 84.0$^{\color{color-}-0.8}$ & 89.6$^{\color{color+}+0.8}$ & 94.4$^{\color{color+}+1.6}$ & 91.6$^{\color{color-}-1.2}$ & 87.6$^{\color{color+}+1.2}$ & 83.2$^{\color{color-}-2.4}$ & 47.6$^{\color{color-}-7.2}$ & 89.2$^{\color{color-}-0.8}$ \\
\textit{late} & \cellcolor{lightred}76.4$^{\color{color-}-8.4}$ & 89.6$^{\color{color+}+0.8}$ & 94.4$^{\color{color+}+1.6}$ & 92.0$^{\color{color-}-0.8}$ & 87.6$^{\color{color+}+1.2}$ & 85.6$^{\color{black}+0.0}$ & 48.4$^{\color{color-}-6.4}$ & 88.8$^{\color{color-}-1.2}$ \\
\midrule
\multicolumn{9}{c}{\textbf{\textit{Target Lang: German (De)}}} \\
\midrule
\textit{Early} & 85.2$^{\color{color+}+0.4}$ & \cellcolor{lightred}85.2$^{\color{color-}-3.6}$ & 94.0$^{\color{color+}+1.2}$ & 90.0$^{\color{color-}-2.8}$ & 87.6$^{\color{color+}+1.2}$ & 83.2$^{\color{color-}-2.4}$ & 48.8$^{\color{color-}-6.0}$ & 89.6$^{\color{color-}-0.4}$ \\
\textit{Middle} & 87.2$^{\color{color+}+2.4}$ & 90.0$^{\color{color+}+1.2}$ & 94.0$^{\color{color+}+1.2}$ & 90.4$^{\color{color-}-2.4}$ & 88.0$^{\color{color+}+1.6}$ & 86.4$^{\color{color+}+0.8}$ & 51.6$^{\color{color-}-3.2}$ & 90.4$^{\color{color+}+0.4}$ \\
\textit{Late} & 85.6$^{\color{color+}+0.8}$ & 88.8$^{\color{black}+0.0}$ & 94.8$^{\color{color+}+2.0}$ & 92.8$^{\color{black}+0.0}$ & 87.2$^{\color{color+}+0.8}$ & 83.6$^{\color{color-}-2.0}$ & 55.6$^{\color{color+}+0.8}$ & 88.0$^{\color{color-}-2.0}$ \\
\midrule
\multicolumn{9}{c}{\textbf{\textit{Target Lang: English (En)}}} \\
\midrule
\textit{Early} & 84.8$^{\color{black}+0.0}$ & 86.8$^{\color{color-}-2.0}$ & 92.0$^{\color{color-}-0.8}$ & 91.6$^{\color{color-}-1.2}$ & 88.0$^{\color{color+}+1.6}$ & 83.6$^{\color{color-}-2.0}$ & 47.6$^{\color{color-}-7.2}$ & 90.4$^{\color{color+}+0.4}$ \\
\textit{Middle} & 86.0$^{\color{color+}+1.2}$ & 88.0$^{\color{color-}-0.8}$ & 92.0$^{\color{color-}-0.8}$ & 91.6$^{\color{color-}-1.2}$ & 88.0$^{\color{color+}+1.6}$ & 85.6$^{\color{black}+0.0}$ & 54.8$^{\color{black}+0.0}$ & 88.0$^{\color{color-}-2.0}$ \\
\textit{Late} & 85.2$^{\color{color+}+0.4}$ & 89.6$^{\color{color+}+0.8}$ & 93.2$^{\color{color+}+0.4}$ & 93.2$^{\color{color+}+0.4}$ & 86.4$^{\color{black}+0.0}$ & 84.8$^{\color{color-}-0.8}$ & 51.2$^{\color{color-}-3.6}$ & 89.2$^{\color{color-}-0.8}$ \\
\midrule
\multicolumn{9}{c}{\textbf{\textit{Target Lang: Spanish (Es)}}} \\
\midrule
\textit{Early} & 84.4$^{\color{color-}-0.4}$ & 89.6$^{\color{color+}+0.8}$ & 95.2$^{\color{color+}+2.4}$ & 92.4$^{\color{color-}-0.4}$ & 88.4$^{\color{color+}+2.0}$ & 82.8$^{\color{color-}-2.8}$ & 48.8$^{\color{color-}-6.0}$ & 88.4$^{\color{color-}-1.6}$ \\
\textit{Middle} & 83.6$^{\color{color-}-1.2}$ & 90.4$^{\color{color+}+1.6}$ & 92.0$^{\color{color-}-0.8}$ & 92.8$^{\color{black}+0.0}$ & 86.8$^{\color{color+}+0.4}$ & 85.6$^{\color{black}+0.0}$ & 53.2$^{\color{color-}-1.6}$ & 89.2$^{\color{color-}-0.8}$ \\
\textit{Late} & 84.0$^{\color{color-}-0.8}$ & 89.6$^{\color{color+}+0.8}$ & 93.6$^{\color{color+}+0.8}$ & 92.8$^{\color{black}+0.0}$ & 87.2$^{\color{color+}+0.8}$ & 85.2$^{\color{color-}-0.4}$ & 52.0$^{\color{color-}-2.8}$ & 89.6$^{\color{color-}-0.4}$ \\
\midrule
\multicolumn{9}{c}{\textbf{\textit{Target Lang: French (Fr)}}} \\
\midrule
\textit{Early} & 86.8$^{\color{color+}+2.0}$ & 88.8$^{\color{black}+0.0}$ & 94.0$^{\color{color+}+1.2}$ & 92.0$^{\color{color-}-0.8}$ & 87.2$^{\color{color+}+0.8}$ & 83.2$^{\color{color-}-2.4}$ & 51.2$^{\color{color-}-3.6}$ & 90.4$^{\color{color+}+0.4}$ \\
\textit{Middle} & 83.2$^{\color{color-}-1.6}$ & 90.0$^{\color{color+}+1.2}$ & 92.8$^{\color{black}+0.0}$ & 90.8$^{\color{color-}-2.0}$ & 86.4$^{\color{black}+0.0}$ & 86.4$^{\color{color+}+0.8}$ & 52.4$^{\color{color-}-2.4}$ & 89.2$^{\color{color-}-0.8}$ \\
\textit{Late} & 83.6$^{\color{color-}-1.2}$ & 88.8$^{\color{black}+0.0}$ & 94.4$^{\color{color+}+1.6}$ & 91.2$^{\color{color-}-1.6}$ & 85.6$^{\color{color-}-0.8}$ & 84.4$^{\color{color-}-1.2}$ & 54.0$^{\color{color-}-0.8}$ & 90.8$^{\color{color+}+0.8}$ \\
\midrule
\multicolumn{9}{c}{\textbf{\textit{Target Lang: Japanese (Ja)}}} \\
\midrule
\textit{Early} & 85.6$^{\color{color+}+0.8}$ & 89.2$^{\color{color+}+0.4}$ & 94.0$^{\color{color+}+1.2}$ & 92.0$^{\color{color-}-0.8}$ & 85.6$^{\color{color-}-0.8}$ & \cellcolor{lightred}82.8$^{\color{color-}-2.8}$ & 52.4$^{\color{color-}-2.4}$ & 88.8$^{\color{color-}-1.2}$ \\
\textit{Middle} & 84.4$^{\color{black}-0.4}$ & 89.6$^{\color{color+}+0.8}$ & 93.6$^{\color{color+}+0.8}$ & 92.4$^{\color{color-}-0.4}$ & 86.4$^{\color{black}+0.0}$ & 86.0$^{\color{color+}+0.4}$ & 52.4$^{\color{color-}-2.4}$ & 89.6$^{\color{color-}-0.4}$ \\
\textit{Late} & 84.4$^{\color{color-}-0.4}$ & 91.2$^{\color{color+}+2.4}$ & 92.8$^{\color{black}+0.0}$ & 92.0$^{\color{color-}-0.8}$ & 87.2$^{\color{color+}+0.8}$ & 86.0$^{\color{color+}+0.4}$ & 52.0$^{\color{color-}-2.8}$ & 90.0$^{\color{black}+0.0}$ \\
\midrule
\multicolumn{9}{c}{\textbf{\textit{Target Lang: Swahili (Sw)}}} \\
\midrule
\textit{Early} & 72.8$^{\color{color-}-12.0}$ & 88.8$^{\color{black}+0.0}$ & 94.4$^{\color{color+}+1.6}$ & 92.4$^{\color{color-}-0.4}$ & 88.8$^{\color{color+}+2.4}$ & 83.6$^{\color{color-}-2.0}$ & \cellcolor{lightred}6.8$^{\color{color-}-48.0}$ & 89.2$^{\color{color-}-0.8}$ \\
\textit{Middle} & 84.8$^{\color{black}+0.0}$ & 90.0$^{\color{color+}+1.2}$ & 93.2$^{\color{color+}+0.4}$ & 91.6$^{\color{color-}-1.2}$ & 88.0$^{\color{color+}+1.6}$ & 86.0$^{\color{color+}+0.4}$ & 48.0$^{\color{color-}-6.8}$ & 89.2$^{\color{color-}-0.8}$ \\
\textit{Late} & 86.0$^{\color{color+}+1.2}$ & 88.8$^{\color{black}+0.0}$ & 93.6$^{\color{color+}+0.8}$ & 91.6$^{\color{color-}-1.2}$ & 86.4$^{\color{black}+0.0}$ & 84.4$^{\color{color-}-1.2}$ & \cellcolor{lightred}30.4$^{\color{color-}-24.4}$ & 89.6$^{\color{color-}-0.4}$ \\
\midrule
\multicolumn{9}{c}{\textbf{\textit{Target Lang: Chinese (Zh)}}} \\
\midrule
\textit{Early} & 84.4$^{\color{color-}-0.4}$ & 89.2$^{\color{color+}+0.4}$ & 93.2$^{\color{color+}+0.4}$ & 90.8$^{\color{color-}-2.0}$ & 87.2$^{\color{color+}+0.8}$ & 84.8$^{\color{color-}-0.8}$ & 47.6$^{\color{color-}-7.2}$ & \cellcolor{lightred}84.4$^{\color{color-}-5.6}$ \\
\textit{Middle} & 85.6$^{\color{color+}+0.8}$ & 90.0$^{\color{color+}+1.2}$ & 94.0$^{\color{color+}+1.2}$ & 91.6$^{\color{color-}-1.2}$ & 87.6$^{\color{color+}+1.2}$ & 84.0$^{\color{color-}-1.6}$ & 54.8$^{\color{black}+0.0}$ & 90.0$^{\color{black}+0.0}$ \\
\textit{Late} & 85.2$^{\color{color+}+0.4}$ & 89.2$^{\color{color+}+0.4}$ & 93.6$^{\color{color+}+0.8}$ & 92.4$^{\color{color-}-0.4}$ & 87.6$^{\color{color+}+1.2}$ & 84.4$^{\color{color-}-1.2}$ & 56.4$^{\color{color+}+1.6}$ & \cellcolor{lightred}87.6$^{\color{color-}-2.4}$ \\
\bottomrule
\end{tabular}
\label{tab:detailed_mgsm_accuracy}
\end{table*}

\begin{table*}[p!]
\centering
\renewcommand{\arraystretch}{1.3}
\setlength{\tabcolsep}{4.5pt}
\caption{Layer-wise intervention effects on language consistency in FLORES-200. We mask language-exclusive experts of target languages at early, middle, and late layers. Significant drops are highlighted in red.}
\small
\begin{tabular}{c|cccccccc}
\toprule
\multirow{2}{*}{\shortstack[c]{\textbf{Intervened} \\ \textbf{Layers}}} & \multicolumn{8}{c}{\textbf{Languages}} \\
 & {Bn} & {De} & {En} & {Es} & {Fr} & {Ja} & {Sw} & {Zh} \\
\midrule
\rowcolor{lightgray} - & 99.6 & 95.7 & 99.7 & 93.6 & 90.8 & 99.6 & 99.1 & 99.4 \\
\midrule
\multicolumn{9}{c}{\textbf{\textit{Target Lang: Bengali (Bn)}}} \\
\midrule
\textit{Early} & 99.9$^{\color{color+}+0.3}$ & 95.9$^{\color{color+}+0.2}$ & 99.7$^{\color{black}+0.0}$ & 93.4$^{\color{color-}-0.2}$ & 90.5$^{\color{color-}-0.3}$ & 99.6$^{\color{black}+0.0}$ & 98.8$^{\color{color-}-0.3}$ & 99.3$^{\color{color-}-0.1}$ \\
\textit{Middle} & 99.6$^{\color{black}+0.0}$ & 96.1$^{\color{color+}+0.4}$ & 99.7$^{\color{black}+0.0}$ & 93.2$^{\color{color-}-0.4}$ & 90.4$^{\color{color-}-0.4}$ & 99.7$^{\color{color+}+0.1}$ & 99.2$^{\color{color+}+0.1}$ & 99.4$^{\color{black}+0.0}$ \\
\textit{Late} & \cellcolor{lightred}46.4$^{\color{color-}-53.2}$ & 95.9$^{\color{color+}+0.2}$ & 99.7$^{\color{black}+0.0}$ & 93.3$^{\color{color-}-0.3}$ & 90.1$^{\color{color-}-0.7}$ & 99.6$^{\color{black}+0.0}$ & 95.9$^{\color{color-}-3.2}$ & 99.4$^{\color{black}+0.0}$ \\
\midrule
\multicolumn{9}{c}{\textbf{\textit{Target Lang: German (De)}}} \\
\midrule
\textit{Early} & 99.6$^{\color{black}+0.0}$ & 95.6$^{\color{color-}-0.1}$ & 99.7$^{\color{black}+0.0}$ & 93.2$^{\color{color-}-0.4}$ & 90.1$^{\color{color-}-0.7}$ & 99.6$^{\color{black}+0.0}$ & 98.9$^{\color{color-}-0.2}$ & 99.4$^{\color{black}+0.0}$ \\
\textit{Middle} & 99.6$^{\color{black}+0.0}$ & 95.9$^{\color{color+}+0.2}$ & 99.7$^{\color{black}+0.0}$ & 93.5$^{\color{color-}-0.1}$ & 90.4$^{\color{color-}-0.4}$ & 99.5$^{\color{color-}-0.1}$ & 99.1$^{\color{black}+0.0}$ & 99.4$^{\color{black}+0.0}$ \\
\textit{Late} & 99.7$^{\color{color+}+0.1}$ & \cellcolor{lightred}21.3$^{\color{color-}-74.4}$ & 99.7$^{\color{black}+0.0}$ & 93.4$^{\color{color-}-0.2}$ & 90.0$^{\color{color-}-0.8}$ & 99.4$^{\color{color-}-0.2}$ & 99.3$^{\color{color+}+0.2}$ & 99.4$^{\color{black}+0.0}$ \\
\midrule
\multicolumn{9}{c}{\textbf{\textit{Target Lang: English (En)}}} \\
\midrule
\textit{Early} & 99.7$^{\color{color+}+0.1}$ & 95.7$^{\color{black}+0.0}$ & 99.7$^{\color{black}+0.0}$ & 93.6$^{\color{black}+0.0}$ & 89.8$^{\color{color-}-1.0}$ & 99.6$^{\color{black}+0.0}$ & 99.0$^{\color{color-}-0.1}$ & 99.4$^{\color{black}+0.0}$ \\
\textit{Middle} & 99.7$^{\color{color+}+0.1}$ & 96.0$^{\color{color+}+0.3}$ & 99.7$^{\color{black}+0.0}$ & 93.2$^{\color{color-}-0.4}$ & 90.5$^{\color{color-}-0.3}$ & 99.6$^{\color{black}+0.0}$ & 99.3$^{\color{color+}+0.2}$ & 99.4$^{\color{black}+0.0}$ \\
\textit{Late} & 99.7$^{\color{color+}+0.1}$ & 95.6$^{\color{color-}-0.1}$ & 99.7$^{\color{black}+0.0}$ & 93.0$^{\color{color-}-0.6}$ & 90.9$^{\color{color+}+0.1}$ & 99.6$^{\color{black}+0.0}$ & 98.8$^{\color{color-}-0.3}$ & 99.3$^{\color{color-}-0.1}$ \\
\midrule
\multicolumn{9}{c}{\textbf{\textit{Target Lang: Spanish (Es)}}} \\
\midrule
\textit{Early} & 99.7$^{\color{color+}+0.1}$ & 95.7$^{\color{black}+0.0}$ & 99.7$^{\color{black}+0.0}$ & 93.4$^{\color{color-}-0.2}$ & 90.3$^{\color{color-}-0.5}$ & 99.6$^{\color{black}+0.0}$ & 99.0$^{\color{color-}-0.1}$ & 99.3$^{\color{color-}-0.1}$ \\
\textit{Middle} & 99.6$^{\color{black}+0.0}$ & 96.1$^{\color{color+}+0.4}$ & 99.7$^{\color{black}+0.0}$ & 93.5$^{\color{color-}-0.1}$ & 90.3$^{\color{color-}-0.5}$ & 99.6$^{\color{black}+0.0}$ & 99.2$^{\color{color+}+0.1}$ & 99.3$^{\color{color-}-0.1}$ \\
\textit{Late} & 99.8$^{\color{color+}+0.2}$ & 95.7$^{\color{black}+0.0}$ & 99.7$^{\color{black}+0.0}$ & \cellcolor{lightred}85.6$^{\color{color-}-8.0}$ & 89.7$^{\color{color-}-1.1}$ & 99.6$^{\color{black}+0.0}$ & 99.3$^{\color{color+}+0.2}$ & 99.3$^{\color{color-}-0.1}$ \\
\midrule
\multicolumn{9}{c}{\textbf{\textit{Target Lang: French (Fr)}}} \\
\midrule
\textit{Early} & 99.8$^{\color{color+}+0.2}$ & 95.7$^{\color{black}+0.0}$ & 99.7$^{\color{black}+0.0}$ & 93.5$^{\color{color-}-0.1}$ & 90.4$^{\color{color-}-0.4}$ & 99.6$^{\color{black}+0.0}$ & 99.0$^{\color{color-}-0.1}$ & 99.4$^{\color{black}+0.0}$ \\
\textit{Middle} & 99.4$^{\color{color-}-0.2}$ & 95.9$^{\color{color+}+0.2}$ & 99.7$^{\color{black}+0.0}$ & 93.4$^{\color{color-}-0.2}$ & 90.2$^{\color{color-}-0.6}$ & 99.6$^{\color{black}+0.0}$ & 99.1$^{\color{black}+0.0}$ & 99.2$^{\color{color-}-0.2}$ \\
\textit{Late} & 99.7$^{\color{color+}+0.1}$ & 96.0$^{\color{color+}+0.3}$ & 99.7$^{\color{black}+0.0}$ & 92.2$^{\color{color-}-1.4}$ & \cellcolor{lightred}13.1$^{\color{color-}-77.6}$ & 99.7$^{\color{color+}+0.1}$ & 98.8$^{\color{color-}-0.3}$ & 99.4$^{\color{black}+0.0}$ \\
\midrule
\multicolumn{9}{c}{\textbf{\textit{Target Lang: Japanese (Ja)}}} \\
\midrule
\textit{Early} & 99.6$^{\color{black}+0.0}$ & 95.7$^{\color{black}+0.0}$ & 99.7$^{\color{black}+0.0}$ & 93.2$^{\color{color-}-0.4}$ & 90.4$^{\color{color-}-0.4}$ & 99.4$^{\color{color-}-0.2}$ & 99.5$^{\color{color+}+0.4}$ & 99.4$^{\color{black}+0.0}$ \\
\textit{Middle} & 99.6$^{\color{black}+0.0}$ & 95.9$^{\color{color+}+0.2}$ & 99.7$^{\color{black}+0.0}$ & 93.6$^{\color{black}+0.0}$ & 90.6$^{\color{color-}-0.2}$ & 99.7$^{\color{color+}+0.1}$ & 98.9$^{\color{color-}-0.2}$ & 99.4$^{\color{black}+0.0}$ \\
\textit{Late} & 99.6$^{\color{black}+0.0}$ & 95.9$^{\color{color+}+0.2}$ & 99.7$^{\color{black}+0.0}$ & 93.6$^{\color{black}+0.0}$ & 90.3$^{\color{color-}-0.5}$ & \cellcolor{lightred}73.7$^{\color{color-}-25.9}$ & 98.9$^{\color{color-}-0.2}$ & 97.4$^{\color{color-}-2.0}$ \\
\midrule
\multicolumn{9}{c}{\textbf{\textit{Target Lang: Swahili (Sw)}}} \\
\midrule
\textit{Early} & 99.6$^{\color{black}+0.0}$ & 95.6$^{\color{color-}-0.1}$ & 99.7$^{\color{black}+0.0}$ & 93.3$^{\color{color-}-0.3}$ & 90.1$^{\color{color-}-0.7}$ & 99.6$^{\color{black}+0.0}$ & 97.5$^{\color{color-}-1.6}$ & 99.4$^{\color{black}+0.0}$ \\
\textit{Middle} & 99.5$^{\color{color-}-0.1}$ & 95.9$^{\color{color+}+0.2}$ & 99.7$^{\color{black}+0.0}$ & 93.3$^{\color{color-}-0.3}$ & 90.2$^{\color{color-}-0.6}$ & 99.6$^{\color{black}+0.0}$ & 98.9$^{\color{color-}-0.2}$ & 99.4$^{\color{black}+0.0}$ \\
\textit{Late} & 99.1$^{\color{color-}-0.5}$ & 95.9$^{\color{color+}+0.2}$ & 99.7$^{\color{black}+0.0}$ & 93.3$^{\color{color-}-0.3}$ & 90.3$^{\color{color-}-0.5}$ & 99.8$^{\color{color+}+0.2}$ & \cellcolor{lightred}82.1$^{\color{color-}-16.9}$ & 99.2$^{\color{color-}-0.2}$ \\
\midrule
\multicolumn{9}{c}{\textbf{\textit{Target Lang: Chinese (Zh)}}} \\
\midrule
\textit{Early} & 99.6$^{\color{black}+0.0}$ & 95.7$^{\color{black}+0.0}$ & 99.6$^{\color{color-}-0.1}$ & 93.3$^{\color{color-}-0.3}$ & 90.2$^{\color{color-}-0.6}$ & 99.5$^{\color{color-}-0.1}$ & 99.0$^{\color{color-}-0.1}$ & 99.3$^{\color{color-}-0.1}$ \\
\textit{Middle} & 99.6$^{\color{black}+0.0}$ & 95.9$^{\color{color+}+0.2}$ & 99.7$^{\color{black}+0.0}$ & 93.4$^{\color{color-}-0.2}$ & 90.5$^{\color{color-}-0.3}$ & 99.7$^{\color{color+}+0.1}$ & 98.9$^{\color{color-}-0.2}$ & 99.5$^{\color{color+}+0.1}$ \\
\textit{Late} & 99.7$^{\color{color+}+0.1}$ & 95.7$^{\color{black}+0.0}$ & 99.7$^{\color{black}+0.0}$ & 93.2$^{\color{color-}-0.4}$ & 90.5$^{\color{color-}-0.3}$ & 98.1$^{\color{color-}-1.5}$ & 99.2$^{\color{color+}+0.1}$ & \cellcolor{lightred}65.7$^{\color{color-}-33.7}$ \\
\bottomrule
\end{tabular}
\label{tab:detailed_flores_consistency}
\end{table*}

\begin{table*}[p!]
\centering
\renewcommand{\arraystretch}{1.3}
\setlength{\tabcolsep}{4.5pt}
\caption{Layer-wise intervention effects on language consistency in MGSM. We mask language-exclusive experts of target languages at early, middle, and late layers. Significant drops are highlighted in red.}
\small
\begin{tabular}{c|cccccccc}
\toprule
\multirow{2}{*}{\shortstack[c]{\textbf{Intervened} \\ \textbf{Layers}}} & \multicolumn{8}{c}{\textbf{Languages}} \\
 & {Bn} & {De} & {En} & {Es} & {Fr} & {Ja} & {Sw} & {Zh} \\
\midrule
\rowcolor{lightgray}
- & 97.6 & 86.0 & 98.8 & 95.6 & 88.0 & 74.8 & 90.8 & 95.2 \\

\midrule
\multicolumn{9}{c}{\textbf{\textit{Target Lang: Bengali (Bn)}}} \\
\midrule
\textit{Early} & 99.2$^{\color{color+}+1.6}$ & 87.2$^{\color{color+}+1.2}$ & 98.4$^{\color{color-}-0.4}$ & 96.0$^{\color{color+}+0.4}$ & 87.2$^{\color{color-}-0.8}$ & 69.2$^{\color{color-}-5.6}$ & 93.2$^{\color{color+}+2.4}$ & 95.6$^{\color{color+}+0.4}$ \\
\textit{Middle} & 98.0$^{\color{color+}+0.4}$ & 88.0$^{\color{color+}+2.0}$ & 99.2$^{\color{color+}+0.4}$ & 95.6$^{\color{black}+0.0}$ & 88.8$^{\color{color+}+0.8}$ & 70.0$^{\color{color-}-4.8}$ & 89.2$^{\color{color-}-1.6}$ & 96.4$^{\color{color+}+1.2}$ \\
\textit{Late} & \cellcolor{lightred}24.8$^{\color{color-}-72.8}$ & 87.2$^{\color{color+}+1.2}$ & 98.4$^{\color{color-}-0.4}$ & 94.4$^{\color{color-}-1.2}$ & 87.2$^{\color{color-}-0.8}$ & 71.6$^{\color{color-}-3.2}$ & 87.6$^{\color{color-}-3.2}$ & 94.4$^{\color{color-}-0.8}$ \\

\midrule
\multicolumn{9}{c}{\textbf{\textit{Target Lang: German (De)}}} \\
\midrule
\textit{Early} & 98.8$^{\color{color+}+1.2}$ & 86.8$^{\color{color+}+0.8}$ & 98.8$^{\color{black}+0.0}$ & 95.2$^{\color{color-}-0.4}$ & 89.2$^{\color{color+}+1.2}$ & 75.2$^{\color{color+}+0.4}$ & 91.2$^{\color{color+}+0.4}$ & 94.8$^{\color{color-}-0.4}$ \\
\textit{Middle} & 98.8$^{\color{color+}+1.2}$ & 86.0$^{\color{black}+0.0}$ & 98.4$^{\color{color-}-0.4}$ & 96.0$^{\color{color+}+0.4}$ & 88.8$^{\color{color+}+0.8}$ & 74.0$^{\color{color-}-0.8}$ & 92.0$^{\color{color+}+1.2}$ & 96.8$^{\color{color+}+1.6}$ \\
\textit{Late} & 98.8$^{\color{color+}+1.2}$ & \cellcolor{lightred}12.4$^{\color{color-}-73.6}$ & 99.6$^{\color{color+}+0.8}$ & 94.4$^{\color{color-}-1.2}$ & 88.0$^{\color{black}+0.0}$ & 68.4$^{\color{color-}-6.4}$ & 80.0$^{\color{color-}-10.8}$ & 68.4$^{\color{color-}-26.8}$ \\

\midrule
\multicolumn{9}{c}{\textbf{\textit{Target Lang: English (En)}}} \\
\midrule
\textit{Early} & 96.0$^{\color{color-}-1.6}$ & 87.6$^{\color{color+}+1.6}$ & 99.2$^{\color{color+}+0.4}$ & 96.0$^{\color{color+}+0.4}$ & 88.8$^{\color{color+}+0.8}$ & 74.8$^{\color{black}+0.0}$ & 97.2$^{\color{color+}+6.4}$ & 96.4$^{\color{color+}+1.2}$ \\
\textit{Middle} & 97.2$^{\color{color-}-0.4}$ & 89.2$^{\color{color+}+3.2}$ & 99.6$^{\color{color+}+0.8}$ & 95.2$^{\color{color-}-0.4}$ & 89.2$^{\color{color+}+1.2}$ & 72.0$^{\color{color-}-2.8}$ & 90.8$^{\color{black}+0.0}$ & 95.6$^{\color{color+}+0.4}$ \\
\textit{Late} & 99.2$^{\color{color+}+1.6}$ & 89.2$^{\color{color+}+3.2}$ & 98.8$^{\color{black}+0.0}$ & 95.6$^{\color{black}+0.0}$ & 88.4$^{\color{color+}+0.4}$ & 73.6$^{\color{color-}-1.2}$ & 90.0$^{\color{color-}-0.8}$ & 96.4$^{\color{color+}+1.2}$ \\

\midrule
\multicolumn{9}{c}{\textbf{\textit{Target Lang: Spanish (Es)}}} \\
\midrule
\textit{Early} & 97.2$^{\color{color-}-0.4}$ & 87.2$^{\color{color+}+1.2}$ & 98.4$^{\color{color-}-0.4}$ & 96.0$^{\color{color+}+0.4}$ & 85.6$^{\color{color-}-2.4}$ & 68.0$^{\color{color-}-6.8}$ & 90.8$^{\color{black}+0.0}$ & 95.2$^{\color{black}+0.0}$ \\
\textit{Middle} & 97.6$^{\color{black}+0.0}$ & 85.6$^{\color{color-}-0.4}$ & 98.8$^{\color{black}+0.0}$ & 94.4$^{\color{color-}-1.2}$ & 86.4$^{\color{color-}-1.6}$ & 70.8$^{\color{color-}-4.0}$ & 90.0$^{\color{color-}-0.8}$ & 94.8$^{\color{color-}-0.4}$ \\
\textit{Late} & 96.4$^{\color{color-}-1.2}$ & 86.4$^{\color{color+}+0.4}$ & 98.8$^{\color{black}+0.0}$ & \cellcolor{lightred}90.4$^{\color{color-}-5.2}$ & 89.6$^{\color{color+}+1.6}$ & 72.8$^{\color{color-}-2.0}$ & 92.8$^{\color{color+}+2.0}$ & 94.8$^{\color{color-}-0.4}$ \\

\midrule
\multicolumn{9}{c}{\textbf{\textit{Target Lang: French (Fr)}}} \\
\midrule
\textit{Early} & 96.0$^{\color{color-}-1.6}$ & 86.0$^{\color{black}+0.0}$ & 98.8$^{\color{black}+0.0}$ & 95.6$^{\color{black}+0.0}$ & 89.6$^{\color{color+}+1.6}$ & 76.8$^{\color{color+}+2.0}$ & 91.6$^{\color{color+}+0.8}$ & 94.0$^{\color{color-}-1.2}$ \\
\textit{Middle} & 97.6$^{\color{black}+0.0}$ & 86.0$^{\color{black}+0.0}$ & 98.8$^{\color{black}+0.0}$ & 94.4$^{\color{color-}-1.2}$ & 88.4$^{\color{color+}+0.4}$ & 73.6$^{\color{color-}-1.2}$ & 92.0$^{\color{color+}+1.2}$ & 96.8$^{\color{color+}+1.6}$ \\
\textit{Late} & 98.4$^{\color{color+}+0.8}$ & 85.2$^{\color{color-}-0.8}$ & 98.4$^{\color{color-}-0.4}$ & 92.0$^{\color{color-}-3.6}$ & \cellcolor{lightred}3.6$^{\color{color-}-84.4}$ & 70.4$^{\color{color-}-4.4}$ & 90.8$^{\color{black}+0.0}$ & 96.0$^{\color{color+}+0.8}$ \\

\midrule
\multicolumn{9}{c}{\textbf{\textit{Target Lang: Japanese (Ja)}}} \\
\midrule
\textit{Early} & 98.4$^{\color{color+}+0.8}$ & 86.8$^{\color{color+}+0.8}$ & 98.4$^{\color{color-}-0.4}$ & 96.0$^{\color{color+}+0.4}$ & 87.6$^{\color{color-}-0.4}$ & 75.2$^{\color{color+}+0.4}$ & 90.8$^{\color{black}+0.0}$ & 95.2$^{\color{black}+0.0}$ \\
\textit{Middle} & 98.0$^{\color{color+}+0.4}$ & 87.6$^{\color{color+}+1.6}$ & 98.4$^{\color{color-}-0.4}$ & 96.0$^{\color{color+}+0.4}$ & 88.8$^{\color{color+}+0.8}$ & 74.4$^{\color{color-}-0.4}$ & 90.4$^{\color{color-}-0.4}$ & 96.4$^{\color{color+}+1.2}$ \\
\textit{Late} & 97.6$^{\color{black}+0.0}$ & 89.6$^{\color{color+}+3.6}$ & 98.8$^{\color{black}+0.0}$ & 96.0$^{\color{color+}+0.4}$ & 87.6$^{\color{color-}-0.4}$ & \cellcolor{lightred}50.8$^{\color{color-}-24.0}$ & 92.0$^{\color{color+}+1.2}$ & 93.6$^{\color{color-}-1.6}$ \\

\midrule
\multicolumn{9}{c}{\textbf{\textit{Target Lang: Swahili (Sw)}}} \\
\midrule
\textit{Early} & 97.6$^{\color{black}+0.0}$ & 86.0$^{\color{black}+0.0}$ & 98.4$^{\color{color-}-0.4}$ & 97.2$^{\color{color+}+1.6}$ & 86.8$^{\color{color-}-1.2}$ & 75.2$^{\color{color+}+0.4}$ & 95.6$^{\color{color+}+4.8}$ & 96.8$^{\color{color+}+1.6}$ \\
\textit{Middle} & 97.6$^{\color{black}+0.0}$ & 87.6$^{\color{color+}+1.6}$ & 98.8$^{\color{black}+0.0}$ & 95.6$^{\color{black}+0.0}$ & 88.4$^{\color{color+}+0.4}$ & 72.0$^{\color{color-}-2.8}$ & 88.8$^{\color{color-}-2.0}$ & 97.2$^{\color{color+}+2.0}$ \\
\textit{Late} & 98.0$^{\color{color+}+0.4}$ & 85.6$^{\color{color-}-0.4}$ & 98.8$^{\color{black}+0.0}$ & 94.8$^{\color{color-}-0.8}$ & 88.8$^{\color{color+}+0.8}$ & 75.2$^{\color{color+}+0.4}$ & \cellcolor{lightred}85.6$^{\color{color-}-5.2}$ & 95.6$^{\color{color+}+0.4}$ \\

\midrule
\multicolumn{9}{c}{\textbf{\textit{Target Lang: Chinese (Zh)}}} \\
\midrule
\textit{Early} & 97.2$^{\color{color-}-0.4}$ & 89.2$^{\color{color+}+3.2}$ & 98.4$^{\color{color-}-0.4}$ & 96.4$^{\color{color+}+0.8}$ & 87.2$^{\color{color-}-0.8}$ & 72.4$^{\color{color-}-2.4}$ & 93.6$^{\color{color+}+2.8}$ & 94.4$^{\color{color-}-0.8}$ \\
\textit{Middle} & 97.2$^{\color{color-}-0.4}$ & 88.4$^{\color{color+}+2.4}$ & 98.4$^{\color{color-}-0.4}$ & 96.0$^{\color{color+}+0.4}$ & 86.4$^{\color{color-}-1.6}$ & 74.8$^{\color{black}+0.0}$ & 89.2$^{\color{color-}-1.6}$ & 95.2$^{\color{black}+0.0}$ \\
\textit{Late} & 98.0$^{\color{color+}+0.4}$ & 85.6$^{\color{color-}-0.4}$ & 98.8$^{\color{black}+0.0}$ & 95.2$^{\color{color-}-0.4}$ & 87.6$^{\color{color-}-0.4}$ & 69.6$^{\color{color-}-5.2}$ & 91.2$^{\color{color+}+0.4}$ & \cellcolor{lightred}89.2$^{\color{color-}-6.0}$ \\

\bottomrule
\end{tabular}
\label{tab:detailed_mgsm_consistency}
\end{table*}

\begin{table*}[t]
\centering
\renewcommand{\arraystretch}{1.3}
\setlength{\tabcolsep}{8pt}
\caption{Layer-wise intervention effects on language consistency in XQuAd. Significant drops are highlighted in red.}
\small
\begin{tabular}{c|ccccc|c}
\toprule
\multirow{2}{*}{\textbf{Intervened Layers}} & \multicolumn{5}{c|}{\textbf{Languages}} & \multirow{2}{*}{\textbf{Overall Avg.}} \\
 & {Ar} & {El} & {En} & {Ru} & {Tr} & \\
\midrule

\rowcolor{lightgray}
- & 63.7 & 62.9 & 67.4 & 54.5 & 58.5 & 61.4 \\

\textit{Early}
& 63.4$^{\color{color-}-0.3}$ 
& 53.9$^{\color{color-}-9.0}$ 
& 62.4$^{\color{color-}-5.0}$ 
& 54.9$^{\color{color+}+0.4}$ 
& 57.1$^{\color{color-}-1.4}$ 
& \cellcolor{lightred}58.3$^{\color{color-}-3.1}$ \\

\textit{Middle}
& 63.4$^{\color{color-}-0.3}$ 
& 64.0$^{\color{color+}+1.1}$ 
& 69.9$^{\color{color+}+2.5}$ 
& 55.2$^{\color{color+}+0.7}$ 
& 56.6$^{\color{color-}-1.9}$ 
& 61.8$^{\color{color+}+0.4}$ \\

\textit{Late}
& 64.3$^{\color{color+}+0.6}$ 
& 63.0$^{\color{color+}+0.1}$ 
& 67.1$^{\color{color-}-0.3}$ 
& 53.8$^{\color{color-}-0.7}$ 
& 56.3$^{\color{color-}-2.2}$ 
& 60.9$^{\color{color-}-0.4}$ \\
\bottomrule
\end{tabular}
\label{tab:xquad_intervention}
\end{table*}

In Section~\ref{sec:intervention}, we primarily focused on the aggregated impact of expert intervention on the target language itself. To provide a more comprehensive evaluation, we present the detailed results in this section. The results for MGSM accuracy are detailed in Table~\ref{tab:detailed_mgsm_accuracy}.
Table~\ref{tab:detailed_flores_consistency} presents the language consistency metrics for the FLORES-200 dataset. The language consistency on MGSM is summarized in Table~\ref{tab:detailed_mgsm_consistency}. The intervention results on XQuAD, serving as a supplementary dataset, are reported in Table~\ref{tab:xquad_intervention}.

\subsection{Routing-Guided Steering}
\label{app:addition-result-steering}

\begin{table*}[t]
\centering
\renewcommand{\arraystretch}{1.3}
\setlength{\tabcolsep}{4pt}
\caption{
Performance comparison of steering at early and late layers on Polymath.
}
\small
\begin{tabular}{c|cccccc|c}
\toprule
\multirow{2}{*}{\textbf{Steering layers}} & \multicolumn{6}{c|}{\textbf{Target Language}} & \multirow{2}{*}{\textbf{Avg}} \\
 & {Bn} & {Sw} & {Fr} & {Es} & {Ja} & {De} & \\
\midrule

\rowcolor{lightgray}
-
& 85.6 & 52.8 & 86.4 & 92.8 & 82.4 & 84.8 & 80.8 \\

\midrule
\textit{Early}
& 85.6$^{\color{black}+0.0}$ 
& 49.6$^{\color{color-}-3.2}$ 
& 86.4$^{\color{black}+0.0}$ 
& 92.8$^{\color{black}+0.0}$ 
& 79.2$^{\color{color-}-3.2}$ 
& 84.0$^{\color{color-}-0.8}$ 
& 79.6$^{\color{color-}-1.2}$ \\

\midrule
\textit{Late}
& 84.0$^{\color{color-}-1.6}$ 
& 47.2$^{\color{color-}-5.6}$ 
& 87.2$^{\color{color+}+0.8}$ 
& 92.8$^{\color{black}+0.0}$ 
& 82.0$^{\color{color+}-0.4}$ 
& 84.8$^{\color{black}+0.0}$ 
& 79.7$^{\color{color-}-1.1}$ \\

\bottomrule
\end{tabular}
\label{tab:layer_steering}
\end{table*}

We further conduct an ablation study by applying routing-guided steering to early and late layers.
As shown in Table~\ref{tab:layer_steering}, aligning routing behavior at early or late layers consistently leads to performance degradation.
This degradation is expected, as steering at these layers disrupts language-specific capabilities that are critical for input understanding and output generation, respectively.

\section{Model Study}

\subsection{Hyperparameter Study}
\label{subsec:hyper}
\begin{figure}[h]
    \centering
    \includegraphics[width=\linewidth]{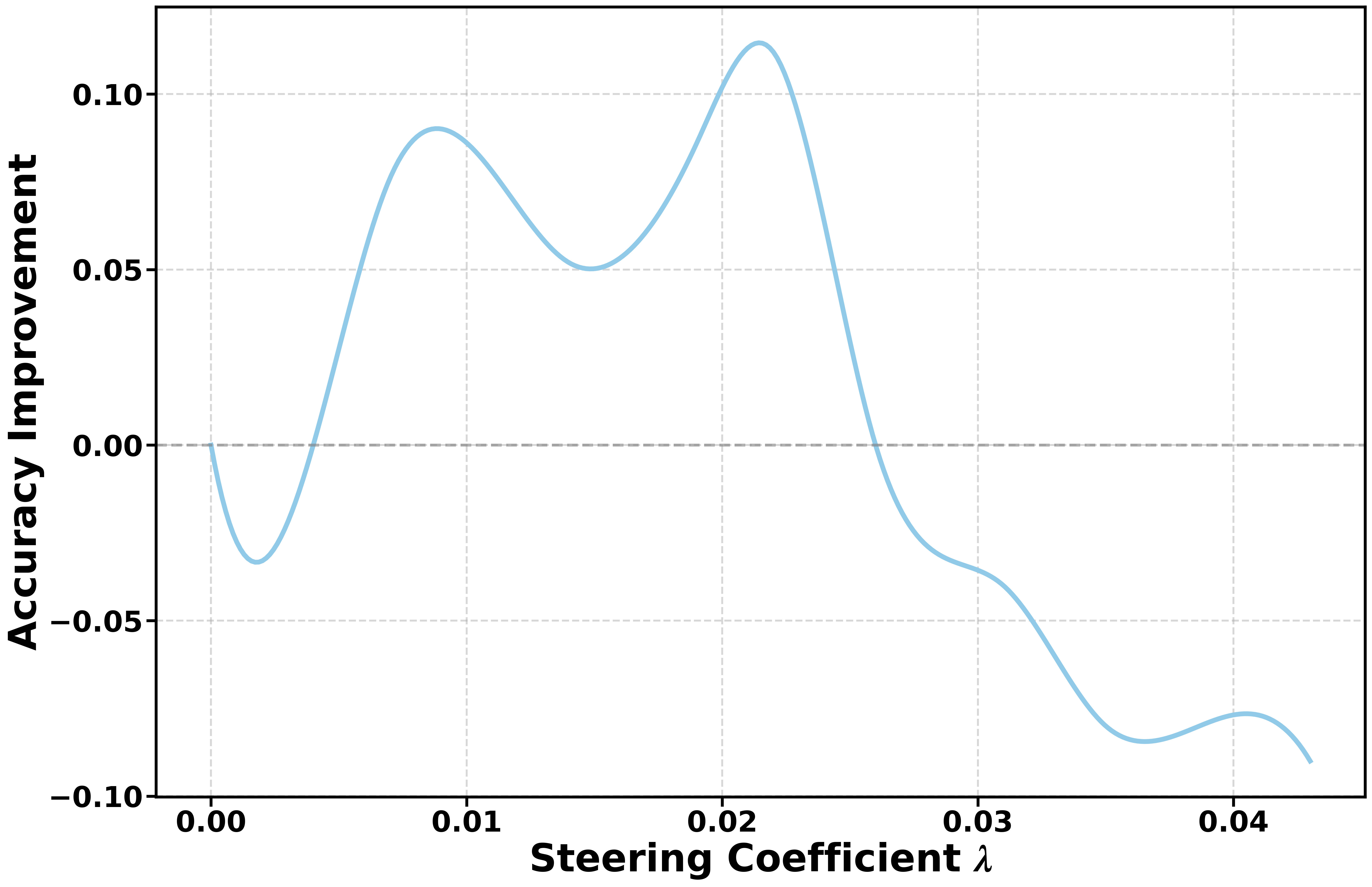}
    \caption{Impact of steering coefficient ($\lambda$) on model performance.}
    \label{fig:total_exclusive}
\end{figure}

We investigate the sensitivity of our proposed method to the steering coefficient, denoted by $\lambda$. This hyperparameter controls the magnitude of the perturbation added to the MoE routing logits, thereby determining the strength with which we bias the expert selection process towards dominant-language experts.

Figure \ref{fig:total_exclusive} illustrates the accuracy improvement across all evaluated target languages as a function of $\lambda$, where the baseline performance is represented by $\lambda=0$. The trajectory of the curve reveals three distinct phases governing the effectiveness of routing steering.

Initially, at minimal values where $\lambda < 0.01$, the steering signal exerts only a marginal influence on the routing distribution. The added perturbation proves insufficient to consistently overcome the innate confidence of the original router, resulting in stochastic fluctuations around the baseline performance rather than meaningful improvement. As $\lambda$ increases into the critical range of $0.01 < \lambda \leq 0.022$, we observe a stable and monotonic increase in cumulative accuracy. In this regime, the steering successfully guides the router toward experts possessing transferable knowledge relevant to the target languages without overwhelming the model's necessary functionalities. The performance culminates in a distinct peak at approximately $\lambda \approx 0.022$, representing the optimal balance for this setup.

Beyond this optimal threshold, specifically when $\lambda > 0.026$, performance degrades precipitously. This sharp decline indicates that an excessively strong steering signal forces a routing distribution that diverges significantly from the model's pre-trained priors. Such over-correction disrupts the delicate balance of expert specialization established during pre-training, leading to a collapse in capability across the target languages. These findings underscore the necessity of carefully tuning $\lambda$ to balance the benefits of guided expert routing against the risk of disrupting the model's foundational representations.

\section{Analysis}
\label{app:Analysis}
This section provides a more detailed analysis of the layer-wise expert intervention results reported in
Table~\ref{tab:intervention_mgsm} and Table~\ref{tab:intervention_mgsm_consistency}.
We focus on two complementary aspects: the emergence of language mixing under late-layer interventions, and the robustness of multilingual capability across languages with different resource levels.

\subsection{Effect of Language Mixing}
Table~\ref{tab:intervention_mgsm_consistency} reveals a clear degradation in language consistency when late-layer language-exclusive experts are masked.
This effect is consistently observed across both MGSM and FLORES-200.
However, as shown in Table~\ref{tab:intervention_mgsm}, task accuracy on MGSM is only mildly affected and occasionally even improves.
This contrast indicates that the model does not fail at task understanding or reasoning.
Instead, the dominant failure mode is a loss of control over the target output language.
Specifically, model outputs frequently switch to dominant or closely related languages under late-layer interventions.
Such language mixing is most pronounced for low-resource languages such as Bn and Sw.
Nevertheless, the same phenomenon also appears in several high-resource languages.
In contrast, English exhibits almost no degradation in language consistency.
This asymmetry suggests that English generation is largely supported by shared or dominant experts.
For non-dominant languages, late-layer language-exclusive experts play a key role in maintaining language identity.
When these experts are removed, the model still completes the task but defaults to its dominant language.
This behavior aligns with prior observations that models tend to mix languages when optimized primarily with outcome-level rewards \cite{Deepseek-R1}.
It further supports the view that multilingual models are inherently dominant-language-centric.
In such cases, enforcing non-dominant languages may introduce a trade-off between task performance and language fidelity.

\subsection{Robustness of Language Capability}
\label{app:robustness}

\begin{table}[t]
\centering
\small
\setlength{\tabcolsep}{6pt} 
\caption{
Layer-wise intervention effects on MGSM averaged by language resource level.
}
\renewcommand{\arraystretch}{1.3}

\begin{tabular}{cccc} 
\toprule
\makecell{\textbf{Intervened} \\ \textbf{Layers}} & \makecell{\textbf{Low-} \\ \textbf{resource}} & \makecell{\textbf{High-} \\ \textbf{resource}} & \textbf{Dominant} \\
\midrule
- & 69.8 & 88.4 & 91.4 \\

\textit{Early} & 
30.2$^{\textcolor{red}{-39.6}}$ & 
87.0$^{\textcolor{red}{-1.4}}$ & 
88.2$^{\textcolor{red}{-3.2}}$ \\

\textit{Middle} & 
66.0$^{\textcolor{red}{-3.8}}$ & 
88.8$^{\textcolor{black}{+0.4}}$ & 
91.0$^{\textcolor{red}{-0.4}}$ \\

\textit{Late} & 
46.4$^{\textcolor{red}{-23.4}}$ & 
88.1$^{\textcolor{red}{-0.3}}$ & 
91.0$^{\textcolor{red}{-0.4}}$ \\

\midrule
\textit{Avg.} & 
47.5$^{\textcolor{red}{-22.3}}$ & 
88.0$^{\textcolor{red}{-0.4}}$ & 
90.1$^{\textcolor{red}{-1.3}}$ \\
\bottomrule
\end{tabular}

\label{tab:intervention_mgsm_resource_avg}
\end{table}

Table~\ref{tab:intervention_mgsm} further highlights substantial differences in robustness across languages with different resource levels.
Low-resource languages exhibit the most severe performance degradation when their language-exclusive experts are masked, especially in early and late layers.
In several cases, accuracy collapses almost entirely, indicating that their multilingual capability relies heavily on a small set of specialized experts.

High-resource languages, by contrast, are significantly more robust.
Although early-layer interventions still cause noticeable drops, the magnitude is much smaller than for low-resource languages, and middle-layer interventions have almost no negative effect.
This suggests that high-resource languages benefit from stronger integration into the shared expert space, allowing them to compensate for the removal of language-exclusive experts by leveraging shared computation paths.

Dominant languages, especially English, exhibit the highest robustness overall.
Even when their own exclusive experts are masked, performance degradation remains limited compared to low-resource languages, and language consistency is largely preserved.
This aligns with our earlier findings that dominant languages are associated with both abundant exclusive experts and strong connectivity to shared experts across layers.

Taken together, these results demonstrate that multilingual capability in MoE models is not uniformly distributed.
Low-resource languages rely heavily on fragile, language-exclusive expert pathways, while high-resource and dominant languages enjoy more resilient representations supported by shared experts.
This robustness gap offers a mechanistic explanation for persistent multilingual performance disparities and further underscores the central role of expert-level specialization in shaping multilingual behavior.


\section{Related Works}
\label{sec:related work}

\subsection{Mechanical Interpretation of Multilingualism}
Prior research on the mechanistic understanding of multilingual behavior in large language models has largely focused on dense architectures.
A number of studies analyze how multilingual representations emerge in dense LLMs by examining internal activations~\citep{activation_1,activation_2}, neurons~\citep{AbstractThought,yiran_howdo}, or attention patterns across languages~\citep{attention_1,attention_2,attention_3}.
Within this line of work, some studies provide evidence that languages engage specific subspaces of models, suggesting non-uniform utilization of model capacity across languages~\citep{tang2024language,kojima2024multilingual,saito2024we,zhang2024unveiling}.
Other studies show that LLMs can also develop concept space that are shared across languages, supporting cross-lingual generalization~\citep{wu2024semantic,wang2024sharing,brinkmann2025large,AbstractThought}.

Despite these insights, mechanistic analyses of multilingualism remain largely confined to dense models.
In contrast, Mixture-of-Experts architectures, which empirically exhibit stronger multilingual performance, introduce explicit expert specialization and dynamic routing, making it crucial to understand how multilingual behavior emerges under such sparse computation.

\subsection{Multilingual Enhancement of LLM}
Enhancing multilingual performance in large language models has long been an active research topic~\citep{zhang2024seallms,dou2025sailor2,Babel,zhou2025disparities,zhang2024enhancing}.
Most existing approaches focus on training-time interventions, improving multilingual capability through data construction and cross-lingual supervision.
A substantial body of work builds high-quality multilingual datasets to expand language coverage and improve robustness across languages~\citep{zhu2024multilingual,she2024mapo,PolyMath,shimabucoro2025post,ko2025understand}.
Other approaches exploit transfer signals from high-resource languages via multilingual fine-tuning or cross-lingual supervision, enabling knowledge and reasoning capabilities to generalize across languages~\citep{zhao2024lens,huo2025enhancing,ruan2025layalign,fan2025slam}.
Beyond training-time methods, recent studies have explored inference-time techniques, such as prompting and steering, to enhance multilingual reasoning without modifying model parameters~\citep{qin2023cross,zhang2024autocap,yong2025crosslingual,gao2025could,tran2025scaling,yu2025cross,son2025linguistic}.

In contrast, we focus on expert routing as a structural enhancement mechanism.
By analyzing and manipulating routing behavior, we study multilingual enhancement from the perspective of expert selection and computation allocation, providing a complementary, architecture-aware view of multilingual improvement.

\section{Use of AI Assistants}
We used AI assistants to support paper writing and code development, primarily for language polishing and programming assistance.
For manuscript preparation, we employed large language models (e.g., GPT-5) to improve sentence fluency and to provide suggestions on clarity and phrasing.
All content generated or suggested by AI assistants was carefully reviewed, verified, and revised by the authors to ensure that the final manuscript faithfully reflects the authors’ intended ideas and technical contributions.
For code development, AI assistants (e.g., GitHub Copilot) were used mainly to generate small code snippets or boilerplate code, which were subsequently adapted, reviewed, and integrated into the overall implementation framework by the authors.
In all cases, the authors retained full control over both the scientific content and the implementation, with AI assistants serving solely as auxiliary tools to improve efficiency rather than as sources of scientific contributions.

\end{document}